\documentclass[letterpaper]{article} 
\usepackage{aaai24}  
\usepackage{times}  
\usepackage{helvet}  
\usepackage{courier}  
\usepackage[hyphens]{url}  
\usepackage{graphicx} 
\usepackage{natbib}  
\usepackage{caption} 
\usepackage{newfloat}
\usepackage{listings}
\DeclareCaptionStyle{ruled}{labelfont=normalfont,labelsep=colon,strut=off} 
\usepackage[ruled,linesnumbered]{algorithm2e}
\usepackage{arydshln}
\usepackage{adjustbox}
\usepackage{amsfonts}       
\usepackage{amsmath}
\usepackage{amssymb}
\usepackage{amsthm}
\usepackage{bm}
\usepackage{dsfont}
\usepackage{booktabs}       
\usepackage{cite}
\usepackage{enumitem} 
\usepackage{graphics}
\usepackage{microtype}      
\usepackage{makecell}
\usepackage{multirow}
\usepackage[subrefformat=parens,labelformat=parens]{subfig}
\usepackage[table]{xcolor} 
\usepackage{xparse}
\usepackage{colortbl}

\usepackage[capitalize]{cleveref}

\newcommand\scalemath[2]{\scalebox{#1}{\mbox{\ensuremath{\displaystyle #2}}}}  

\usepackage{amsthm}

\allowdisplaybreaks

\title{Turning Waste into Wealth: Leveraging Low-Quality Samples for Enhancing Continuous Conditional Generative Adversarial Networks}
\author{
	Xin Ding\textsuperscript{\rm 1},
	Yongwei Wang\textsuperscript{\rm 2}\thanks{Corresponding author.},
	Zuheng Xu\textsuperscript{\rm 3}
}
\affiliations{
	\textsuperscript{\rm 1} School of Artificial Intelligence, Nanjing University of Information Science \& Technology, Nanjing, China \\
	\textsuperscript{\rm 2} Shanghai Institute for Advanced Study, Zhejiang University, Shanghai, China \\
	\textsuperscript{\rm 3} Department of Statistics, University of British Columbia, Vancouver, BC, Canada \\
	dingxin@nuist.edu.cn, yongwei.wang@zju.edu.cn, zuheng.xu@stat.ubc.ca
}

\graphicspath{{images/}} 

\begin{document}

\maketitle

\begin{abstract}
	Continuous Conditional Generative Adversarial Networks (CcGANs) enable generative modeling conditional on continuous scalar variables (termed regression labels). However, they can produce subpar fake images due to limited training data. Although Negative Data Augmentation (NDA) effectively enhances unconditional and class-conditional GANs by introducing anomalies into real training images, guiding the GANs away from low-quality outputs, its impact on CcGANs is limited, as it fails to replicate negative samples that may occur during the CcGAN sampling. We present a novel NDA approach called Dual-NDA specifically tailored for CcGANs to address this problem. Dual-NDA employs two types of negative samples: visually unrealistic images generated from a pre-trained CcGAN and label-inconsistent images created by manipulating real images' labels. Leveraging these negative samples, we introduce a novel discriminator objective alongside a modified CcGAN training algorithm. Empirical analysis on UTKFace and Steering Angle reveals that Dual-NDA consistently enhances the visual fidelity and label consistency of fake images generated by CcGANs, exhibiting a substantial performance gain over the vanilla NDA. Moreover, by applying Dual-NDA, CcGANs demonstrate a remarkable advancement beyond the capabilities of state-of-the-art conditional GANs and diffusion models, establishing a new pinnacle of performance. Our codes can be found at \url{https://github.com/UBCDingXin/Dual-NDA}.
\end{abstract}

\section{Introduction}

The objective of Continuous Conditional Generative Modeling (CCGM), as illustrated in \Cref{fig:illustrative_CCGM}, is to estimate the distribution of high-dimensional data, such as images, in terms of continuous scalar variables, referred to as regression labels. However, this task is very challenging due to insufficient or even zero training images for certain regression labels and the absence of a suitable label input mechanism. 

In a recent breakthrough, \citet{ding2021ccgan, ding2023ccgan} introduced the pioneering model for this purpose, termed Continuous Conditional Generative Adversarial Networks (CcGANs), showcasing their superiority over conventional conditional GANs across various regression datasets. CcGANs have a wide spectrum of practical applications, including engineering inverse design~\citep{heyrani2021pcdgan, fang2023diverse}, medical image analysis~\citep{dufumier2021contrastive}, remote sensing image analysis~\citep{giry2022sar}, model compression~\citep{ding2023distilling,shi2023regression}, point cloud generation~\citep{triess2022point}, carbon sequestration~\citep{stepien2023continuous}, data-driven solutions for poroelasticity~\citep{kadeethum2022continuous}, etc. However, it's important to note that while CcGANs have shown success in these tasks, challenges remain when dealing with extremely sparse or imbalanced training data, leaving ample room to improve CcGAN models further.

To alleviate the data sparsity or imbalance issue, traditional data augmentation techniques for GANs \citep{zhao2020differentiable, karras2020training, tran2021data, jiang2021deceive, tseng2021regularizing, liu2022combating} often employ geometric transformations on real images, such as flipping, translation, and rotation, to guide GANs in learning ``what to generate". However, \citet{sinha2021negative} introduced a distinctive approach, known as Negative Data Augmentation (NDA), for unconditional or class-conditional GANs. The approach, depicted in \Cref{fig:example_NDA}, intentionally crafts negative samples via transformations applied to training images, encompassing techniques like Jigsaw~\citep{noroozi2016unsupervised}, Stitching, Mixup~\citep{zhang2018mixup}, Cutout~\citep{devries2017improved}, and CutMix~\citep{yun2019cutmix}. These negative samples, akin to those produced by the generator, are presented as fake images (in contrast to real images from the training set) and incorporated into the discriminator's training, instructing GANs on ``what to avoid". Nevertheless, NDA's application is limited in the context of CcGANs, as it cannot replicate negative samples that may arise from pre-trained CcGANs, illustrated by two types of representative low-quality images in \Cref{fig:example_ccgan_bad_fake_images}.

\begin{figure*}[!htbp] 
	\centering
	\includegraphics[width=0.9\linewidth]{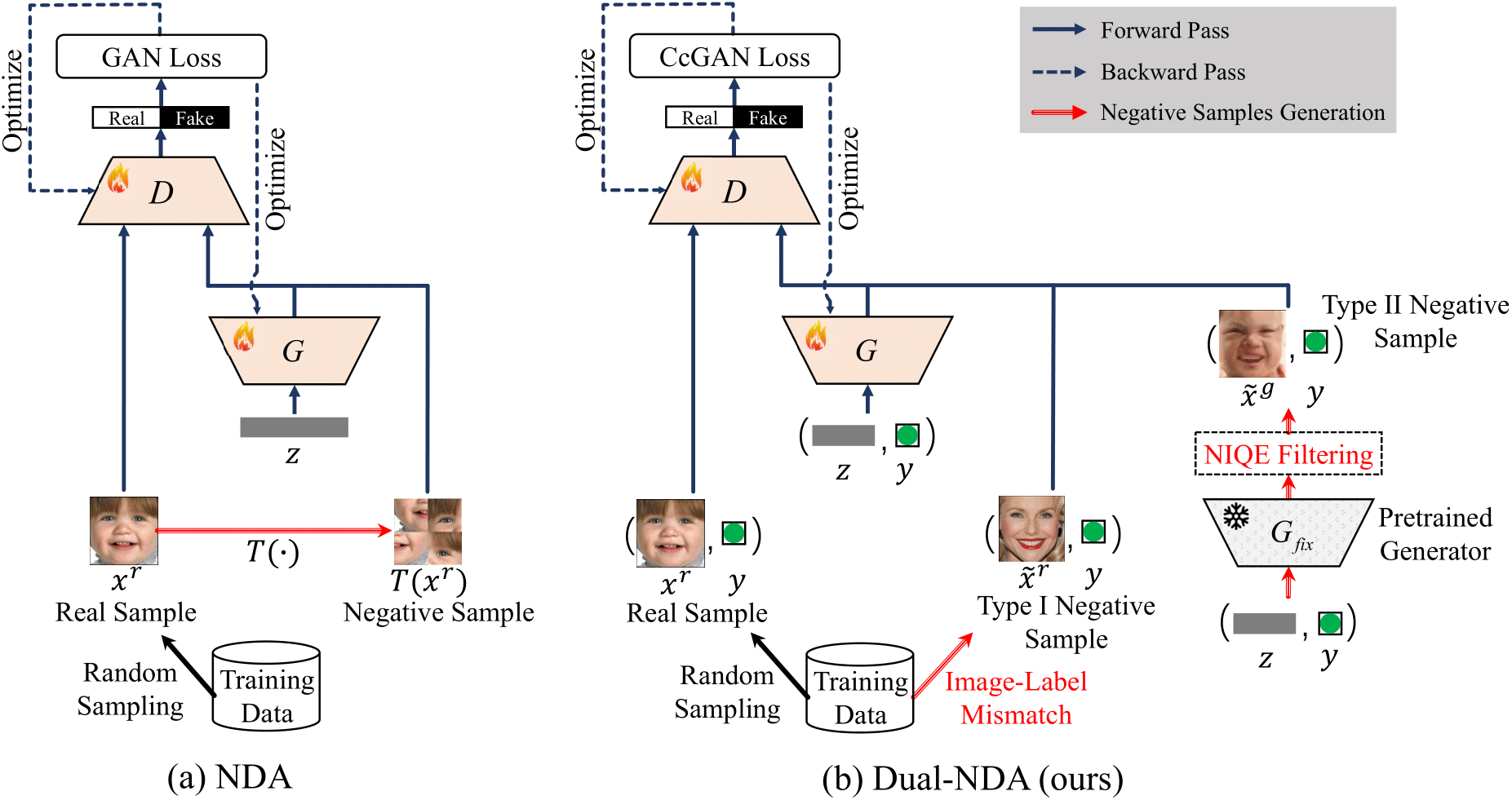}
	\caption{Illustrative workflows for the vanilla NDA~\citep{sinha2021negative} and our proposed Dual-NDA. }
	\label{fig:illustrative_workflow}
\end{figure*}

\begin{figure}[!htbp] 
	\centering
	\includegraphics[width=1\linewidth]{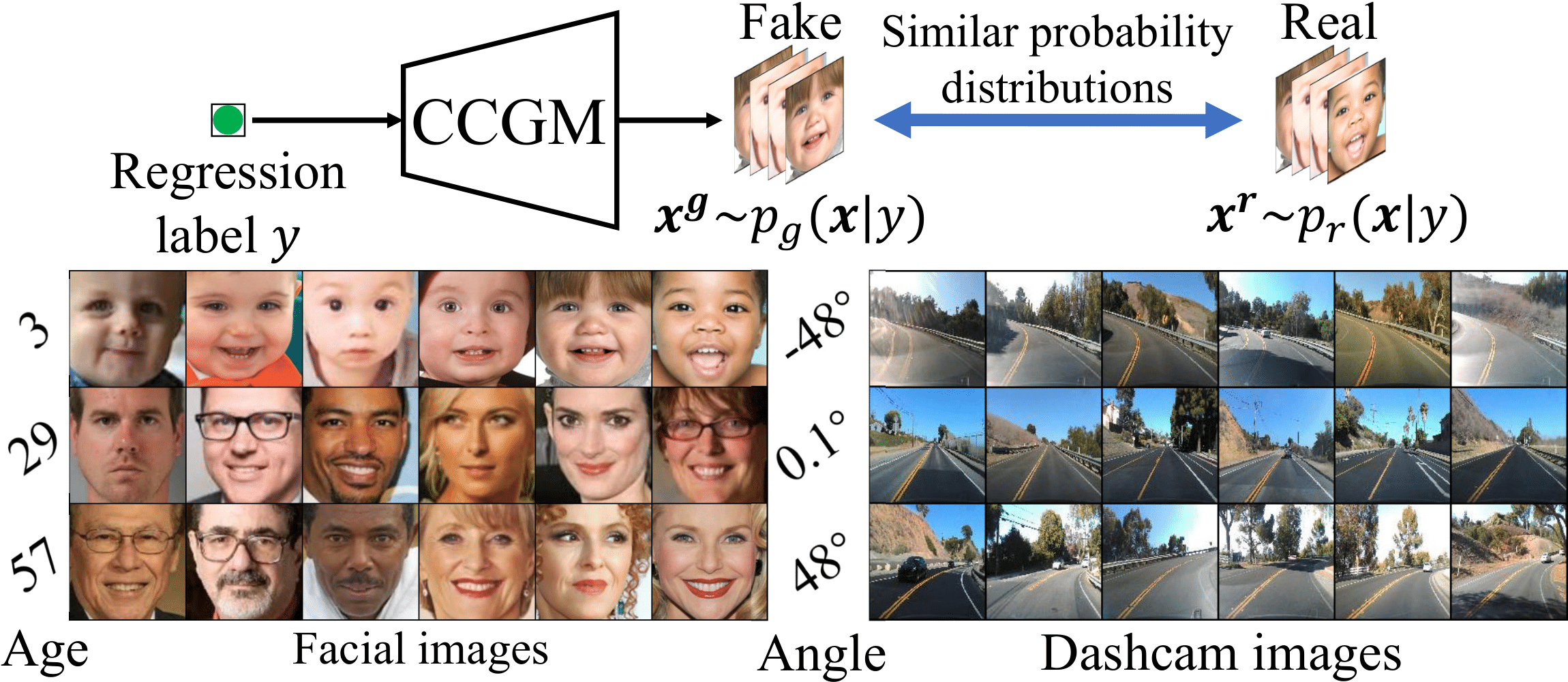}
	\caption{Illustration of the CCGM task and sample images from two regression datasets (UTKFace and Steering Angle). }
	\label{fig:illustrative_CCGM}
\end{figure}

\begin{figure}[!htbp] 
	\centering
	\includegraphics[width=1\linewidth]{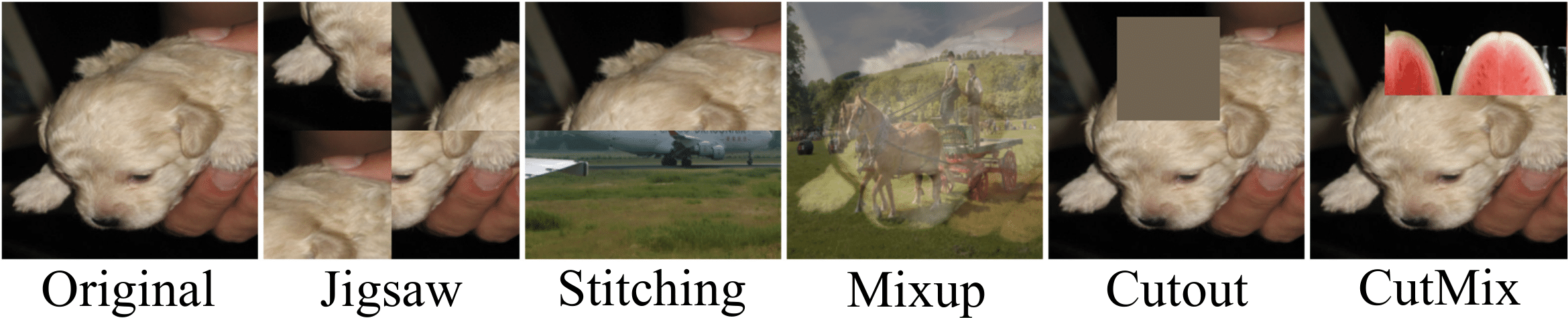}
	\caption{Example negative samples from NDA by transforming a realistic training image (from \citet{sinha2021negative}).}
	\label{fig:example_NDA}
\end{figure}

\begin{figure}[!htbp] 
	\centering
	\includegraphics[width=0.9\linewidth]{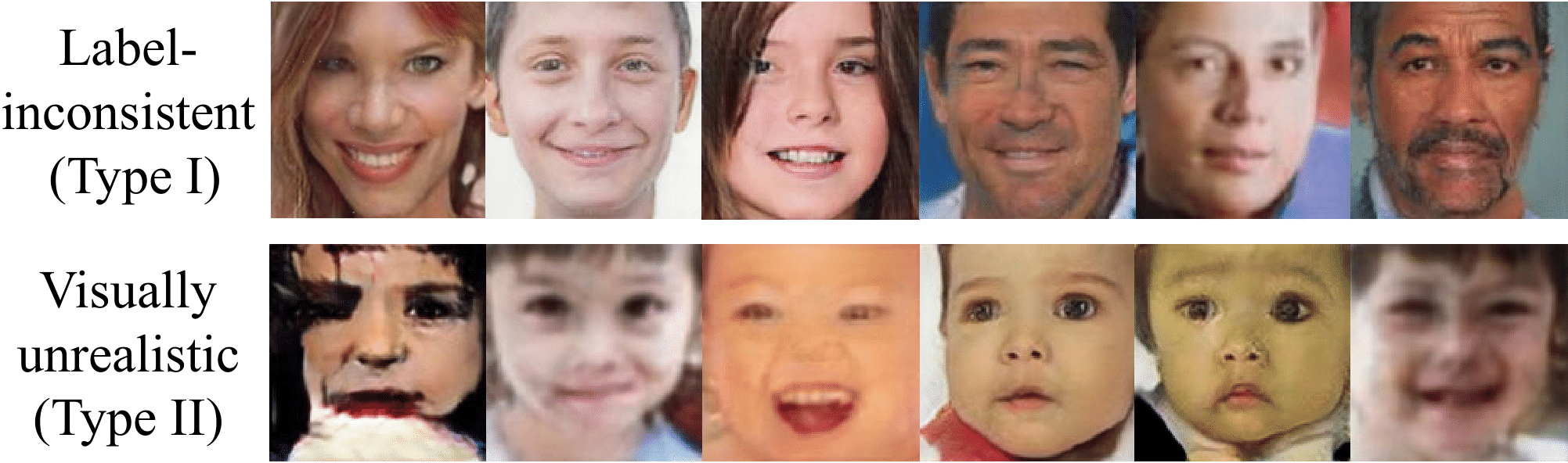}
	\caption{Some actual low-quality fake images generated from a pre-trained CcGAN at ``Age=3" on UTKFace. The term ``label-inconsistent" indicates that these fake images do not align with the conditioning label.}\label{fig:example_ccgan_bad_fake_images}
\end{figure}

To address this challenge, we introduce a novel NDA strategy termed Dual-NDA in this work, as depicted in \Cref{fig:illustrative_workflow}. Unlike the synthetic images showcased in \Cref{fig:example_NDA}, Dual-NDA enriches the training set of CcGANs with two categories of negative samples, strategically mirroring the low-quality images seen in \Cref{fig:example_ccgan_bad_fake_images}. Type I negative samples are mismatched image-label pairs formed by manipulating labels of real images in the training set. In contrast, Type II negative samples are generated by evaluating fake images from a pre-trained generator and retaining those exhibiting the poorest visual quality. Based on these two types of negative samples, we devise a new soft vicinal discriminator loss tailored to enhance the training of CcGANs. Our comprehensive experimental investigation demonstrates the effectiveness of these negative samples in improving CcGAN performance, particularly in terms of visual quality and label consistency enhancement. Notably, CcGANs can generally achieve remarkable superiority through Dual-NDA over state-of-the-art class-conditional GANs and diffusion models.

Our major contributions can be summarized as follows: (1) We propose Dual-NDA, a novel NDA strategy tailored specifically for CcGANs. (2) We present a novel Dual-NDA-based framework for CcGAN training, incorporating a new vicinal discriminator loss. (3) Through extensive experiments, we substantiate the efficacy of Dual-NDA, demonstrating its positive impact on CcGANs' performance. (4) Our work extends beyond existing literature \citep{ding2021ccgan, ding2023ccgan} by providing a comprehensive comparative analysis, benchmarking CcGAN-based approaches against cutting-edge conditional generative models.

\section{Related Work}

\subsection{Continuous Conditional GANs}

Conditional Generative Adversarial Networks (cGANs) (cGANs), introduced by \citet{mirza2014conditional}, extend the vanilla GAN~\citep{goodfellow2014generative} to cater to the Conditional Generative Modeling (CGM) scenario. In this context, a condition denoted as $y$ is provided as input to both the generator and discriminator networks. Mathematically, cGANs are designed to estimate the density function $p_r(\bm{x}|y)$ characterizing the underlying conditional data distribution, with the generator network tasked with approximating this distribution through an estimated density function denoted as $p_g(\bm{x}|y)$. While conventional cGANs, as explored in various works~\citep{brock2018large, zhang2019self, kang2021rebooting, hou2022conditional}, mainly focus on scenarios involving discrete conditions like class labels or text descriptions, their applicability faces limitations when dealing with CGM tasks involving regression labels as conditions. Such limitations arise primarily due to the scarcity of available real images for certain regression labels and the challenge of encoding regression labels that may take infinitely many values.

To address these challenges, \citet{ding2021ccgan, ding2023ccgan} propose the Continuous Conditional GAN (CcGAN) framework, which incorporates novel empirical cGAN losses and label input mechanisms. To mitigate data insufficiency, CcGANs leverage real images within a hard/soft vicinity of $y$ to estimate $p_r(\bm{x}|y)$. This leads to the definition of the Hard/Soft Vicinal Discriminator Loss (HVDL/SVDL) and the generator loss for CcGANs as follows:
\begin{align}
	& \widehat{\mathcal{L}}(D)= \nonumber\\
	& - \frac{1}{N^r}\sum_{j=1}^{N^r}\sum_{i=1}^{N^r}\mathds{E}_{\epsilon^r\sim\mathcal{N}(0,\sigma^2)}\left[W_1\log(D(\bm{x}_i^r, y_j^r+\epsilon^r)) \right] \nonumber\\
	&- \frac{1}{N^g}\sum_{j=1}^{N^g}\sum_{i=1}^{N^g}\mathds{E}_{\epsilon^g\sim\mathcal{N}(0,\sigma^2)}\left[ W_2\log(1-D(\bm{x}_i^g, y_j^g+\epsilon^g)) \right], \label{eq:vicinal_disc_loss} \\
	& \widehat{\mathcal{L}}(G) =  \nonumber\\
	& - \frac{1}{N^g}\sum_{i=1}^{N^g}\mathds{E}_{\epsilon^g\sim\mathcal{N}(0,\sigma^2)}\log (D(G(\bm{z}_{i}, y_i^g+\epsilon^g), y_i^g+\epsilon^g)), \label{eq:gen_loss}
\end{align}
where $D(\bm{x},y)$ and $G(\bm{z},y)$ represent the discriminator and generator networks, $\bm{x}_i^r$ and $\bm{x}_i^g$ are real and fake images, $y_i^r$ and $y_i^g$ are real and fake labels, $\bm{z}_i$ is Gaussian noise, $N^r$ and $N^g$ are sample sizes, $\sigma$ is a hyperparameter controlling the variance of Gaussian noise $\epsilon$, and the weights $W_1$ and $W_2$ are determined by the types of vicinity. For HVDL with the hard vicinity, $W_1$ and $W_2$ are defined as follows:
\begin{equation}
	\label{eq:weight_hard}
	\scalemath{1}{
		W_1=\frac{\mathds{1}_{\{|y_j^r+\epsilon-y_i^r|\leq\kappa\}}}{N_{y^r_j+\epsilon,\kappa}^r},\quad W_2=\frac{\mathds{1}_{\{|y_j^g+\epsilon-y_i^g|\leq\kappa\}}}{N_{y^g_j+\epsilon,\kappa}^g},
	}
\end{equation}
where $\mathds{1}$ is an indicator function, $N_{y,\kappa}$ is the number of the $y_i$'s satisfying $|y-y_i|\leq\kappa$. For SVDL with the soft vicinity, $W_1$ and $W_2$ are defined as:
\begin{equation}
	\label{eq:weight_soft}
	\scalemath{1}{
		W_1=\frac{w(y_i^r,y_j^r+\epsilon)}{\sum_{i=1}^{N^r}w(y_i^r,y_j^r+\epsilon)}, \quad  W_2=\frac{w(y_i^g,y_j^g+\epsilon)}{\sum_{i=1}^{N^g}w(y_i^g,y_j^g+\epsilon)},
	}
\end{equation}
where $w(y, y^\prime) = e^{-\nu(y-y^\prime)^2}$. The hyperparameters $\kappa$, $\nu$ and $\sigma$ in the above formulae can be determined using the rule of thumb in \citet{ding2021ccgan, ding2023ccgan}. It is worth noting that both \cref{eq:vicinal_disc_loss} and \cref{eq:gen_loss} are derived by reformulating the vanilla cGAN loss~\citep{mirza2014conditional}. Similarly, \citet{ding2023ccgan} also reformulated the GAN hinge loss~\citep{lim2017geometric} for the CcGAN training. Furthermore, to address the challenge of encoding regression labels, \citet{ding2021ccgan, ding2023ccgan} developed the Naive Label Input (NLI) and Improved Label Input (ILI) mechanisms.  The effectiveness of CcGANs has been substantiated across diverse datasets.

\subsection{Negative Data Augmentation}

To enhance unconditional and class-conditional GANs, \citet{sinha2021negative} introduced a novel Negative Data Augmentation (NDA) strategy that intentionally creates out-of-distribution negative samples by applying various transformations, such as Jigsaw, Stitching, and Mixup, to realistic training images. Furthermore, \citet{sinha2021negative} introduced a new GAN training objective incorporating these negative samples. This training objective guides the generator network away from producing low-quality outputs resembling the negative samples, thereby encouraging the generator to create more realistic and diverse images. An illustrative workflow of NDA is provided in \Cref{fig:illustrative_workflow}. NDA has demonstrated its effectiveness in enhancing the performance of BigGAN~\citep{brock2018large}, a popular cGAN model, across both unconditional and class-conditional generative modeling tasks, highlighting NDA's potential for advancing GAN-driven image synthesis. However, as demonstrated in \Cref{fig:example_NDA} and \Cref{fig:example_ccgan_bad_fake_images}, NDA-generated negative samples markedly differ from the low-quality samples produced by CcGANs, which explains the limited impact of NDA in our experimental study. Consequently, we propose a novel NDA strategy named Dual-NDA, aimed at generating the two distinct types of negative samples illustrated in \Cref{fig:example_ccgan_bad_fake_images}.

\section{Proposed Method: Dual-NDA}

\subsection{Overview}

We present Dual-NDA, an innovative NDA strategy tailored to enhance the performance of CcGANs. The comprehensive workflow of Dual-NDA is depicted in \Cref{fig:illustrative_workflow}. This approach introduces a novel training paradigm for CcGANs, comprising two key components: a carefully designed negative sample generation mechanism adept at imitating the low-quality images shown in \Cref{fig:example_ccgan_bad_fake_images}, and a new vicinal discriminator loss that harnesses these negative samples to enhance the visual quality and label consistency of generated images.

\subsection{Two Types of Negative Samples}

Dual-NDA generates two types of negative samples, denoted as Type I and Type II, respectively. Some example negative samples are presented in Appendix C.

\vspace{0.2em}
{\setlength{\parindent}{0cm} \textbf{(1) Type I: Label-Inconsistent Real Images}} 
\vspace{0.2em}

{\setlength{\parindent}{0cm} Type I negative samples comprise label-inconsistent real images generated through the dynamic mismatching of image-label pairs during the discriminator training. Recall the CcGAN training for $D(\bm{x},y)$. When estimating the image distribution conditional on a selected label $y \in [0,1]$, denoted as $p_r(\bm{x}|y)$, both real and fake images with labels falling within a hard or soft vicinity of $y$ are selected to train the discriminator $D(\bm{x},y)$. However, using a very wide vicinity could potentially lead to label-inconsistency issues. To solve this problem, Dual-NDA generates Type I negative samples labeled by $y$ through the following steps:
\begin{enumerate}[label=\alph*)]
	\item Compute the pairwise absolute distances between $y$ and the labels of all real images in the training set, denoted as $d_i^y=|y-y_i^r|$, $i=1,...,N^r$. These absolute distance values form an array, denoted as $\bm{d}^y=[d_1^y,d_2^y,...,d_{N^r}^y]$.
	\item Subsequently, calculate the $q_1$-th quantile of $\bm{d}^y$, denoted as $c_{q_1}^y$, with $q_1$ being a hyperparameter typically set in the range of $0.3$ to $0.9$.
	\item Choose training images with labels greater than $c_{q_1}^y$ to form the Type I negative samples for label $y$. It's important to note that the actual labels of these selected real images are significantly different from $y$. 
\end{enumerate}
Based on this mechanism, and considering the observed $y$'s in the CcGAN training process, a set of all Type I negative samples can be constructed, represented as follows:
\begin{equation}
	\label{eq:set_type_1}
	\begin{aligned}
		Q_\text{I} = & \{(\tilde{\bm{x}}_i^{r}, y_i^\text{I})| \tilde{\bm{x}}_i^{r}\sim p_r(\bm{x}|y_i^\text{I}), y_i^\text{I}\sim p(y), \\ & |\tilde{y}_i^{r}-y_i^\text{I}|>c_{q_1}^{y_i^\text{I}},  i=1,...,N^r_\text{I} \},
	\end{aligned}
\end{equation}
where $p_r(\bm{x}|y_i^\text{I})$ denotes the density of real images' distribution conditional on $y_i^\text{I}$, $p(y)$ is the density of the labels' distribution, $\tilde{y}_i^{r}$ is the actual label of $\tilde{\bm{x}}_i^{r}$, and $N^r_\text{I}$ is the sample size. Please note that this process for generating Type I negative samples is seamlessly integrated into the training algorithm of CcGANs, which is outlined in detail in Algorithm 3 provided in Appendix A.

\vspace{0.2em}
{\setlength{\parindent}{0cm} \textbf{(2) Type II: Visually Unrealistic Fake Images}}
\vspace{0.2em}

{\setlength{\parindent}{0cm} Type II negative samples are visually unrealistic fake images generated using a combination of a frozen CcGAN generator and a NIQE filtering mechanism. The generation process involves the following procedures:
\begin{enumerate}[label=\alph*)]
	\itemsep-0.05em 
	\item Dual-NDA initiates the process by sampling a large number of fake image-label pairs from the frozen CcGAN generator, denoted as $Q^g=\{(\bm{x}_i^g,y_i^g)| i=1,...,M\}$.
	
	\item Then, the NIQE filtering mechanism assesses the visual quality of fake images using the Naturalness Image Quality Evaluator (NIQE)~\citep{mittal2012making}, where NIQE is a popular metric used to gauge the quality of an image based on its realism and natural appearance. The corresponding NIQE model has been pretrained on the training set for CcGANs. 
	
	\item In accordance with the NIQE metric, the filtering mechanism selects fake images that possess NIQE scores surpassing a predetermined threshold, effectively targeting images with the poorest visual quality. 
\end{enumerate}
The fake image-label pairs that are chosen through this process constitute the Type II negative samples, denoted as
\begin{equation}
	\label{eq:set_type_2}
	\scalemath{1}{
		\begin{aligned}
			Q_\text{II} = \{ (\tilde{\bm{x}}_i^{g}, \tilde{y}_i^{g})|&  \text{NIQE}(\tilde{\bm{x}}_i^{g})>c_{q_2}, (\tilde{\bm{x}}_i^{g}, \tilde{y}_i^{g})\in Q^g, \\
			& i=1,...,N^g_\text{II}, N^g_\text{II}<M \}, 
		\end{aligned}
	}
\end{equation}
where $c_{q_2}$ is a threshold determined by a hyperparameter $q_2$ and may have a connection with $\tilde{y}_i^{g}$. Dual-NDA typically works well if $q_2$ locates in $[0.5,0.9]$. It's important to note that we adopt two distinct strategies to determine $c_{q_2}$: 
\begin{itemize}
	\itemsep-0.05em 
	\item For labels with integer-valued observations (e.g., ages or counting numbers), a separate $c_{q_2}$ value is computed for each distinct integer value of the regression label. Specifically, assuming the regression label of interest has $K$ distinct integer values, denoted as $\{ 1,..., k ..., K \}$, we denote the corresponding fake images labeled by $k$ as $\bm{X}^g_{k}=\{\bm{x}_{1}^{g,k},..., \bm{x}_{M_{k}}^{g,k} \}$. Then, we compute the NIQE scores of images in $\bm{X}_{k}$, and store the scores in an array, denoted as $\bm{s}^k=[s_1^k,s_2^k,...,s_{M_{k}}^k]$. Subsequently, we calculate the $q_2$-th quantile of $\bm{s}^k$, denoted as $c^k_{q_2}$. Finally, we choose fake images in $\bm{X}_{k}$ with NIQE scores larger than $c^k_{q_2}$. The above procedures are executed for each of the $K$ distinct integer values and summarized in Algorithm 1 in Appendix A.
	
	\item For strictly continuous labels (e.g., angles), a single global threshold $c_{q_2}$ is chosen instead of using $K$ separate values. In this case, $c_{q_2}$ is the $q_2$-th quantile of the NIQE scores of all fake images in $Q^g$, and fake images with NIQE scores larger than $c_{q_2}$ are kept as Type II negative samples. We summarize this process in Algorithm 2 in Appendix A.
\end{itemize}

\subsection{A Modified Training Mechanism for CcGANs}

With the Type I and Type II negative samples separated respectively into $Q_\text{I}$ and $Q_\text{II}$, we introduce a modified training mechanism for CcGANs, illustrated in-depth in Algorithm 3 in Appendix A. The core of this training mechanism is an innovative vicinal discriminator loss, delineated as follows:
\begin{equation}
	\label{eq:new_vicinal_disc_loss}
	\scalemath{1}{
		\begin{aligned}
			& \widetilde{\mathcal{L}}(D)= \\
			& - \frac{1}{N^r}\sum_{j=1}^{N^r}\sum_{i=1}^{N^r}\mathds{E}_{\epsilon\sim\mathcal{N}(0,\sigma^2)}\left[W_1\log(D(\bm{x}_i^r, y_j^r+\epsilon)) \right] \\
			&- \frac{1-\bar{\lambda}}{N^g}\sum_{j=1}^{N^g}\sum_{i=1}^{N^g}\mathds{E}_{\epsilon\sim\mathcal{N}(0,\sigma^2)}\left[ W_2\log(1-D(\bm{x}_i^g, y_j^g+\epsilon)) \right] \\
			&-\frac{\lambda_1}{N_\text{I}}\sum_{i=1}^{N_\text{I}}\log(1-D(\tilde{\bm{x}}_i^{r}, y_i^\text{I}))) \\
			&-\frac{\lambda_2}{N_\text{II}}\sum_{j=1}^{N_\text{II}}\sum_{i=1}^{N_\text{II}}\mathds{E}_{\epsilon\sim\mathcal{N}(0,\sigma^2)}\left[ W_3\log\left(1-D\left(\tilde{\bm{x}}_i^{g}, \tilde{y}_j^{g}+\epsilon\right)\right) \right], 
		\end{aligned}
	}
\end{equation}
where $\lambda_1$ and $\lambda_2$ are two hyperparameters with values in $[0,1]$, $\bar{\lambda}=\lambda_1+\lambda_2$, $(\tilde{\bm{x}}_i^{r}, y_i^\text{I})\in Q_\text{I}$, $(\tilde{\bm{x}}_i^{g}, \tilde{y}_i^{g})\in Q_\text{II}$. While the weights $W_1$ and $W_2$ have been defined in \cref{eq:weight_soft}, $W_3$ aligns with \cref{eq:weight_hard}, taking the form:
\begin{equation}
	\label{eq:w3_for_niqe}
	\scalemath{0.95}{
		W_3=\frac{\mathds{1}_{\{|\tilde{y}_j^{g}+\epsilon-\tilde{y}_i^{g}|\leq\kappa\}}}{N_{\tilde{y}^{g}_j+\epsilon,\kappa}^g}.
	}
\end{equation}
Within \cref{eq:new_vicinal_disc_loss}, the first two terms stem from the SVDL presented in \cref{eq:vicinal_disc_loss}. The third and fourth terms are grounded in the Type I and Type II negative samples, respectively. The hyperparameters $\lambda_1$ and $\lambda_2$ control the influence of negative samples on the CcGAN training. Such an effect has been carefully studied in our experiment. Furthermore, the hyperparameters $\kappa$, $\nu$, and $\sigma$ in \cref{eq:new_vicinal_disc_loss} are determined using the empirically validated guidelines in literature~\citep{ding2023ccgan}. It is important to note that, in the new training mechanism, the generator loss is consistent with \cref{eq:gen_loss}. Furthermore, an alternative version of \cref{eq:new_vicinal_disc_loss} based on hinge loss is also provided in Appendix A.

\section{Experiments}

\subsection{Experimental Setup}

{\setlength{\parindent}{0cm}\textbf{Datasets.}} Following \citet{ding2023ccgan}, we utilize two regression datasets, namely UTKFace~\citep{utkface} and Steering Angle~\citep{steeringangle}, for our experimental study. The UTKFace dataset includes human facial images annotated with ages for regression analysis. We use the pre-processed version \cite{ding2023ccgan}, which consists of 14,723 RGB images. Age labels span from 1 to 60, showcasing a broad age distribution with varying sample sizes per age group (ranging from 50 to 1051 images). The Steering Angle dataset, derived from a subset of an autonomous driving dataset \citep{steeringangle}, comprises 12,271 RGB images. These images are annotated with 1,774 distinct steering angles, spanning from $-80.0^\circ$ to $80.0^\circ$. All Steering Angle images are captured using a dashboard camera on a car, with simultaneous recording of the corresponding steering wheel rotation angles. Both datasets offer two versions, encompassing resolutions of $64\times 64$ and $128\times 128$, respectively, resulting in a total of four datasets. Illustrative training images are visualized in \Cref{fig:illustrative_CCGM}, while data distributions are presented in Appendix B.

{\setlength{\parindent}{0cm}\textbf{Baseline Methods.}} To substantiate the effectiveness of Dual-NDA, we choose the following state-of-the-art conditional generative models as baselines for comparison: (1) Two class-conditional GANs are chosen in the comparison, including ReACGAN~\citep{kang2021rebooting} and ADCGAN~\citep{hou2022conditional}. (2) Two cutting-edge class-conditional diffusion models, including classifier guidance (ADM-G)~\citep{dhariwal2021diffusion} and classifier-free guidance (CFG)~\citep{ho2021classifier} models, are also included in the comparison. (3) Three CcGAN-based methods are compared, consisting of CcGAN (SVDL+ILI) w/o NDA~\citep{ding2023ccgan}, CcGAN w/ NDA~\citep{sinha2021negative}, and the proposed Dual-NDA. 

{\setlength{\parindent}{0cm}\textbf{Training Setup.}} In the implementation of class-conditional GANs and diffusion models, we bin the regression labels of UTKFace into 60 classes and those of Steering Angle into 221 classes. For low-resolution experiments, we re-implement the baseline CcGANs, while for high-resolution experiments, we utilize the checkpoints provided by \citet{ding2023ccgan}. To implement NDA, we set the hyperparameter $\lambda$ to 0.25, adhering to the setup in \citet{sinha2021negative}. \textit{For Dual-NDA, we utilize the pre-trained generator of ``CcGAN w/o NDA" in conjunction with the NIQE filtering process to generate Type II negative samples.} We successfully generate 60,000 and 17,740 Type II negative samples, respectively, in the UTKFace and Steering Angle experiments. Regarding Dual-NDA's hyperparameters, in UTKFace experiments, $\lambda_1$ is set to 0.05 and $\lambda_2$ to 0.15. In Steering Angle experiments at 64$\times$64 resolution, $\lambda_1$ is set to 0.1 and $\lambda_2$ to 0.2, while at 128$\times$128 resolution, $\lambda_1$ is set to 0.2 and $\lambda_2$ to 0.3. Additionally, we set $q_1$ to 0.9 for UTKFace and 0.5 for Steering Angle, with $q_2$ consistently set to 0.9. More detailed training setups can be found in Appendix B.

{\setlength{\parindent}{0cm}\textbf{Evaluation Setup.}} Following \citet{ding2023ccgan}, we generate 60,000 fake images for the UTKFace experiments and 100,000 fake images for the Steering Angle experiments from each compared method. These generated images are subject to evaluation using both an overall metric and three separate metrics. Specifically, the Sliding Fr\'echet Inception Distance (SFID)~\citep{ding2023ccgan} is taken as the overall metric. Notably, the radius utilized for SFID computations is set to 0 for the UTKFace experiments and 2 for the Steering Angle experiments. Furthermore,  NIQE~\citep{mittal2012making}, Diversity~\citep{ding2023ccgan}, and Label Score~\citep{ding2023ccgan} metrics are employed to respectively gauge visual fidelity, diversity, and label consistency of the generated images.

\subsection{Experimental Results}

We present a performance comparison of various methods on the four datasets in \Cref{tab:main_results}, complemented by illustrative figures presented in \Cref{fig:line_graph_UK128} and \Cref{fig:line_graph_SA128}. Analysis of the table and figures reveal the following findings:
\begin{itemize}

	\item Among the assessed methods, Dual-NDA demonstrates superior performance across all four datasets in terms of SFID and NIQE metrics. Its distinct advantage becomes more pronounced on the Steering Angle datasets, thereby underscoring its effectiveness in tackling more intricate CCGM scenarios.
	
	\item Compared with the baseline ``CcGAN w/o NDA", Dual-NDA exhibits reduced values in terms of NIQE and Label Score metrics. This trend is further supported by \Cref{fig:line_graph_UK128} and \Cref{fig:line_graph_SA128}, which illustrate a substantial reduction in NIQE and Label Score values across the entire range of regression labels. These outcomes suggest that incorporating Type I and II negative samples effectively enhances the visual quality and label consistency of the generated samples.
	
	\item NDA consistently fails to yield desirable effects, and in all cases, it leads to a decline in the performance of CcGANs.
	
	\item Class-conditional GANs and diffusion models often exhibit limited effectiveness. Particularly when applied to the Steering Angle datasets, most of them suffer from the mode collapse problem. This outcome once more underscores the prevailing superiority of CcGANs over class-conditional generative models.
	
	\item A crucial observation to highlight is that certain methods, like ADCGAN, might exhibit higher Diversity scores than Dual-NDA. However, it's important to note that many of their diverse generated images might be label-inconsistent, consequently leading to high Label Score values.
\end{itemize}

\begin{table*}[!h] 
	\centering
	\begin{adjustbox}{width=1\textwidth}
		\begin{tabular}{ccccccc}
			\toprule
			\textbf{Dataset} & \textbf{Type} & \textbf{Method} & \begin{tabular}[c]{@{}c@{}} \textbf{SFID} $\downarrow$ \\ (overall quality)\end{tabular}   & \begin{tabular}[c]{@{}c@{}} \textbf{NIQE} $\downarrow$ \\ (visual fidelity)\end{tabular} & \begin{tabular}[c]{@{}c@{}} \textbf{Diversity} $\uparrow$ \\ {(diversity)}\end{tabular} & \begin{tabular}[c]{@{}c@{}} \textbf{Label Score} $\downarrow$ \\ (label consistency)\end{tabular} \\
			\midrule
			
			\multirow{7}[0]{*}{\begin{tabular}[c]{@{}c@{}} \textbf{UTKFace} \\ {(64$\times$64)}\end{tabular}} & \multirow{4}[0]{*}{\begin{tabular}[c]{@{}c@{}} Class-conditional\\ {(60 classes)}\end{tabular}} & ReACGAN (Neurips'21) & 0.548 (0.136) & \underline{1.679 (0.170)} & 1.206 (0.240) & \underline{6.846 (5.954)} \\
			& & ADCGAN (ICML'22) & 0.573 (0.218) & 1.680 (0.140) & \textbf{1.451 (0.019)} & 17.574 (12.388) \\
			& & ADM-G (Neurips'21) & 0.744 (0.195) & 2.856 (0.225) & 0.917 (0.318) & 7.583 (6.066) \\
			& & CFG (Neurips'21) & 2.155 (0.638) & 1.681 (0.303) & 0.858 (0.413) & 8.477 (7.820) \\ \cdashline{2-7}
			
			& \multirow{3}[0]{*}{\begin{tabular}[c]{@{}c@{}} CcGAN \\ (SVDL+ILI)\end{tabular} } & w/o NDA (T-PAMI'23) & \underline{0.413 (0.155)} & 1.733 (0.189) & 1.329 (0.161) & 8.240 (6.271) \\
			& & w/ NDA (ICLR'21) & 0.491 (0.157) & 1.757 (0.183) & \underline{1.399 (0.130)} & 8.229 (6.713) \\
			& & \textbf{Dual-NDA (ours)} & \textbf{0.396 (0.153)} & \textbf{1.678 (0.183)} & 1.298 (0.187) & \textbf{6.765 (5.600)} \\ \midrule
			
			\multirow{7}[0]{*}{\begin{tabular}[c]{@{}c@{}} \textbf{UTKFace} \\ {(128$\times$128)}\end{tabular}} &\multirow{4}[0]{*}{\begin{tabular}[c]{@{}c@{}} Class-conditional\\ {(60 classes)}\end{tabular}} & ReACGAN (Neurips'21) & 0.445 (0.098) & 1.426 (0.064) & 1.152 (0.304) & \textbf{6.005} (5.182) \\
			& & ADCGAN (ICML'22) & 0.468 (0.143) & 1.231 (0.048) & \textbf{1.365 (0.035)} & 15.777 (11.572) \\ 
			& & ADM-G (Neurips'21) &  0.997 (0.208)  & 3.705 (0.409) & 0.831 (0.271) & 11.618 (8.754) \\
			& & CFG (Neurips'21) & 1.521 (0.333) & 1.888 (0.263) & 1.170 (0.174) & 11.430 (9.917) \\ \cdashline{2-7}
			
			& \multirow{3}[0]{*}{\begin{tabular}[c]{@{}c@{}} CcGAN \\ (SVDL+ILI)\end{tabular}} & w/o NDA (T-PAMI'23) & \underline{0.367 (0.123)} & \underline{1.113 (0.033)} & 1.199 (0.232) & 7.747 (6.580) \\
			& & w/ NDA (ICLR'21) & 1.136 (0.244) & 1.125 (0.049) & 0.986 (0.471) & 6.384 (5.324) \\
			& & \textbf{Dual-NDA (ours)} & \textbf{0.361 (0.127)} & \textbf{1.081 (0.042)} & \underline{1.257 (0.238)} & \underline{6.310 (5.194)} \\ \midrule

			\multirow{7}[0]{*}{\begin{tabular}[c]{@{}c@{}} \textbf{Steering Angle} \\ {(64$\times$64)}\end{tabular}} & \multirow{4}[0]{*}{\begin{tabular}[c]{@{}c@{}} Class-conditional\\ {(221 classes)}\end{tabular}} & ReACGAN (Neurips'21) & 3.635 (0.491) & 2.099 (0.072) & 0.543 (0.366) & 27.277 (21.508) \\
			& & ADCGAN (ICML'22) & 2.960 (1.083) & 2.015 (0.003) & 0.930 (0.018) & 40.535 (24.031) \\
			& & ADM-G (Neurips'21) &  2.890 (0.547)  &  2.164 (0.200)  &  0.205 (0.160) &  24.186 (20.685) \\
			& & CFG (Neurips'21) & 4.703 (0.894) & 2.070 (0.022) & 0.923 (0.119) & 56.663 (39.914) \\ \cdashline{2-7}
			
			& \multirow{3}[0]{*}{\begin{tabular}[c]{@{}c@{}} CcGAN \\ (SVDL+ILI)\end{tabular}} & w/o NDA (T-PAMI'23) & \underline{1.334 (0.531)} & \underline{1.784 (0.065)} & \underline{1.234 (0.209)} & 14.807 (14.297) \\
			& & w/ NDA (ICLR'21) & 1.381 (0.527) & 1.994 (0.081) & 1.231 (0.167) & \textbf{10.717 (10.371)} \\
			& & \textbf{Dual-NDA (ours)} & \textbf{1.114 (0.503)} & \textbf{1.738 (0.055)} & \textbf{1.251 (0.172)} & \underline{11.809 (11.694)} \\ \midrule
			
			\multirow{7}[0]{*}{\begin{tabular}[c]{@{}c@{}} \textbf{Steering Angle} \\ {(128$\times$128)}\end{tabular}} &\multirow{4}[0]{*}{\begin{tabular}[c]{@{}c@{}} Class-conditional\\ {(221 classes)}\end{tabular}} & ReACGAN (Neurips'21) & 3.979 (0.919) & 4.060 (0.643) & 0.250 (0.269) & 36.631 (38.592) \\
			& & ADCGAN (ICML'22) & 3.110 (0.799) & 5.181 (0.010) & 0.001 (0.001) & 44.242 (29.223) \\ 
			& & ADM-G (Neurips'21) &  \underline{1.593 (0.449)}  & 3.476 (0.153)  & \underline{1.120 (0.121)} & 32.040 (27.836) \\
			& & CFG (Neurips'21) & 5.425 (1.573) & 2.742 (0.109) & 0.762 (0.121) & 50.015 (34.640) \\ \cdashline{2-7}
			
			& \multirow{3}[0]{*}{\begin{tabular}[c]{@{}c@{}} CcGAN \\ (SVDL+ILI)\end{tabular}} & w/o NDA (T-PAMI'23) & 1.689 (0.443) & \underline{2.411 (0.100)} & 1.088 (0.243) & 18.438 (16.072) \\
			& & w/ NDA (ICLR'21) & 1.736 (0.562) & 2.435 (0.160) & 1.022 (0.247) & \textbf{12.438 (11.612)} \\
			& & \textbf{Dual-NDA (ours)} & \textbf{1.390 (0.421)} & \textbf{2.135 (0.065)} & \textbf{1.133 (0.217)} & \underline{14.099 (12.097)} \\
			\bottomrule
		\end{tabular}%
	\end{adjustbox}
	\caption{Average quality of fake images from compared methods with standard deviations in the parentheses. ``$\downarrow$" (``$\uparrow$") indicates lower (higher) values are preferred. The best and second-best results are marked respectively in boldface and underlined. We re-implement all compared methods except for ``CcGAN w/o NDA" in the $128\times 128$ experiments.}
	\label{tab:main_results}%
\end{table*}%

\begin{figure}[!h] 
	\centering
	\includegraphics[width=1.0\linewidth]{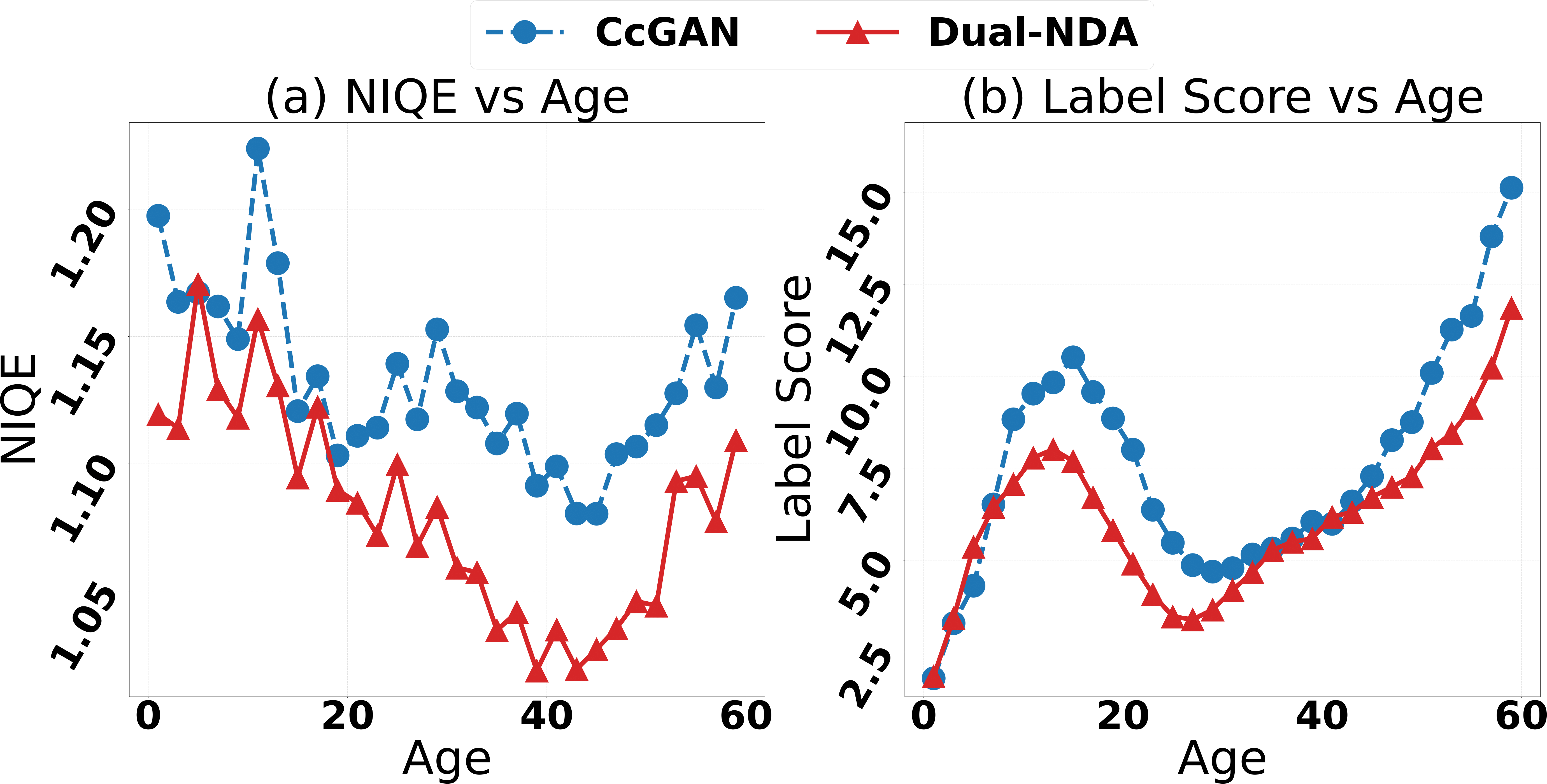}
	\caption{Line graphs of NIQE/Label Score versus Age for the UTKFace (128$\times$128) experiment.}\label{fig:line_graph_UK128}
\end{figure}

\begin{figure}[!h] 
	\centering
	\includegraphics[width=1.0\linewidth]{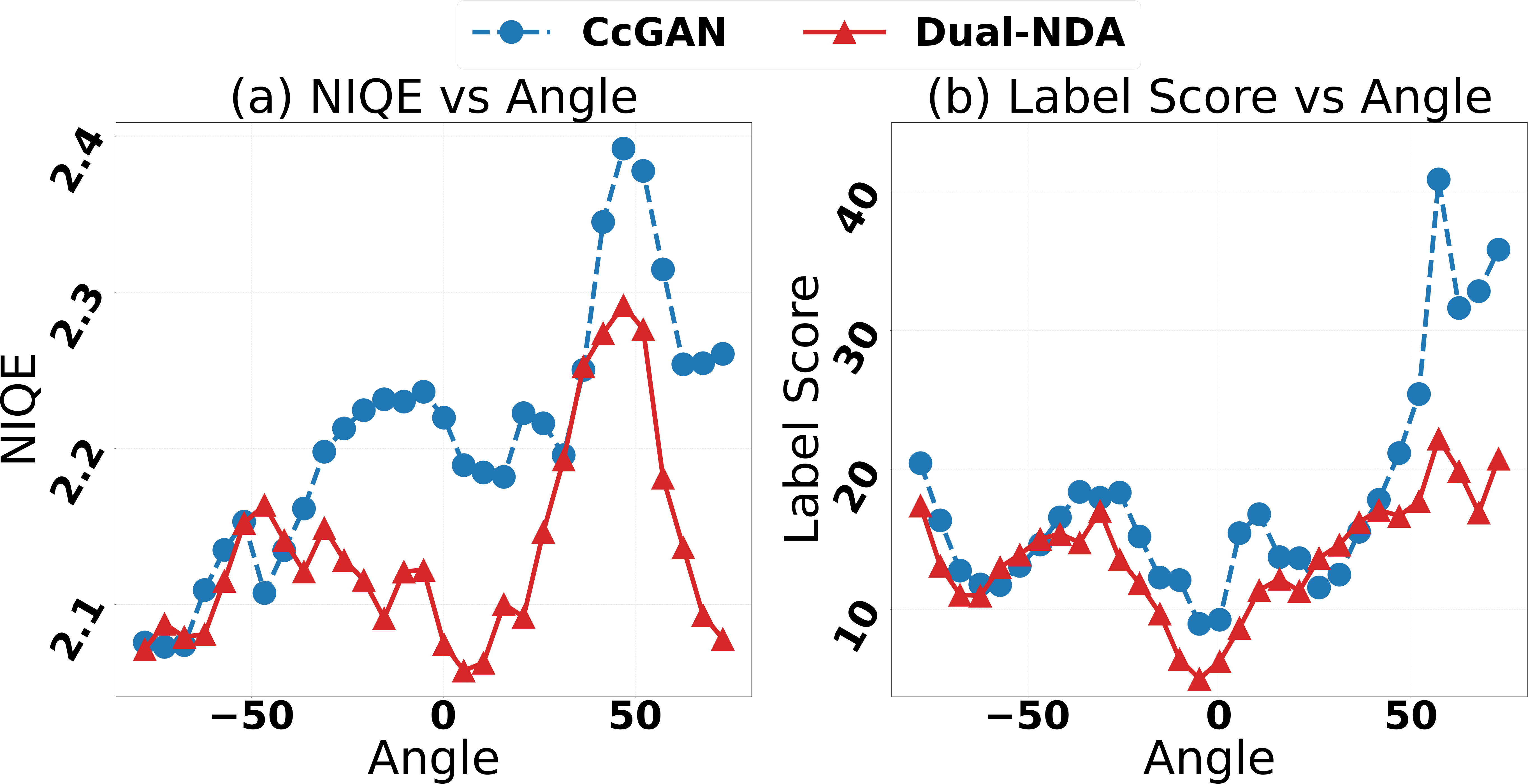}
	\caption{Line graphs of NIQE/Label Score versus Angle for the Steering Angle (128$\times$128) experiment.}\label{fig:line_graph_SA128}
\end{figure}

\subsection{Ablation Study}

We also conduct comprehensive ablation studies on the Steering Angle (64$\times$64) dataset. These studies are designed to systematically investigate the impact of individual components and key hyperparameters within the framework of Dual-NDA, as outlined below. The standard deviations for the evaluation scores are provided in Appendix C.
\begin{enumerate}[label=\alph*)]
	\item \textbf{The Effects of Type I and II Negative Samples}: We individually delve into the distinct impacts of Type I and Type II negative samples. The experimental results are summarized in \Cref{fig:SA64_ablation_effect_types}. From this visual result, we can deduce that the integration of Type I negative samples during training effectively enhances the label consistency of generated images (evidenced by decreased Label Score values), yet it might potentially lead to a compromise in visual quality (reflected by higher NIQE values relative to the blue dashed line). Interestingly, the influence of Type II negative samples exhibits a unique pattern. These samples, designed to enhance visual quality, surprisingly contribute to an improvement in label consistency (as reflected by a Label Score value below the red dashed line). This phenomenon can be rationalized by the insights provided by \citet{ding2023distilling}, which suggest that certain label-inconsistent fake images often exhibit poor visual quality. Consequently, the removal of visually unrealistic images may lead to an enhancement in label consistency. Combining both Type I and Type II negative samples ultimately leads to the most favorable overall performance.
	
	\item \textbf{The Combined Impact of $\lambda_1$ and $\lambda_2$}: Instead of examining the individual effects of $\lambda_1$ and $\lambda_2$, we delve into their combined impact on the SFID scores of CcGANs. We achieve this by varying $\lambda_1+\lambda_2$ within the range of 0.2 to 0.7. The experimental outcomes, shown in \Cref{fig:SA64_ablation_effect_lambda}, highlight a consistent trend: Dual-NDA consistently outperforms the baseline CcGAN across a wide spectrum of $\lambda_1+\lambda_2$ values in terms of SFID. Notably, adjustments to $\lambda_1+\lambda_2$ within the interval of $[0.2,0.5]$ often yield optimal performance for CcGANs.
	
	\item \textbf{The Effect of $q_1$}: We also experiment to analyze the impact of $q_1$ on Dual-NDA's performance. We set a grid of values for $q_1$, and the corresponding quantitative result is presented in \Cref{tab:effect_of_q1}. These findings show that Dual-NDA's performance is not significantly affected by variations in $q_1$. Empirically, $q_1$ with values such as 0.5 or 0.9 often produces favorable outcomes.
	
	\item \textbf{The Effect of $q_2$}: In the last ablation study, we examine the influence of $q_2$ on Dual-NDA's performance and report the quantitative results in \Cref{tab:effect_of_q2}. These results reveal that Dual-NDA is insensitive to the values of $q_2$, so we let $q_2=0.9$ throughout our experiments.
	
	\item \textbf{Creating Type II negative samples from other generators}: As indicated in Table~\ref{tab:different_generators}, we incorporate generators from ADCGAN and ReACGAN to generate Type II negative samples. Nonetheless, it is noteworthy that both ADCGAN and ReACGAN exhibit an adverse impact on the NIQE values, indicating a decrease in visual quality.
	
\end{enumerate}

\begin{figure}[!h] 
	\centering
	\includegraphics[width=0.8\linewidth]{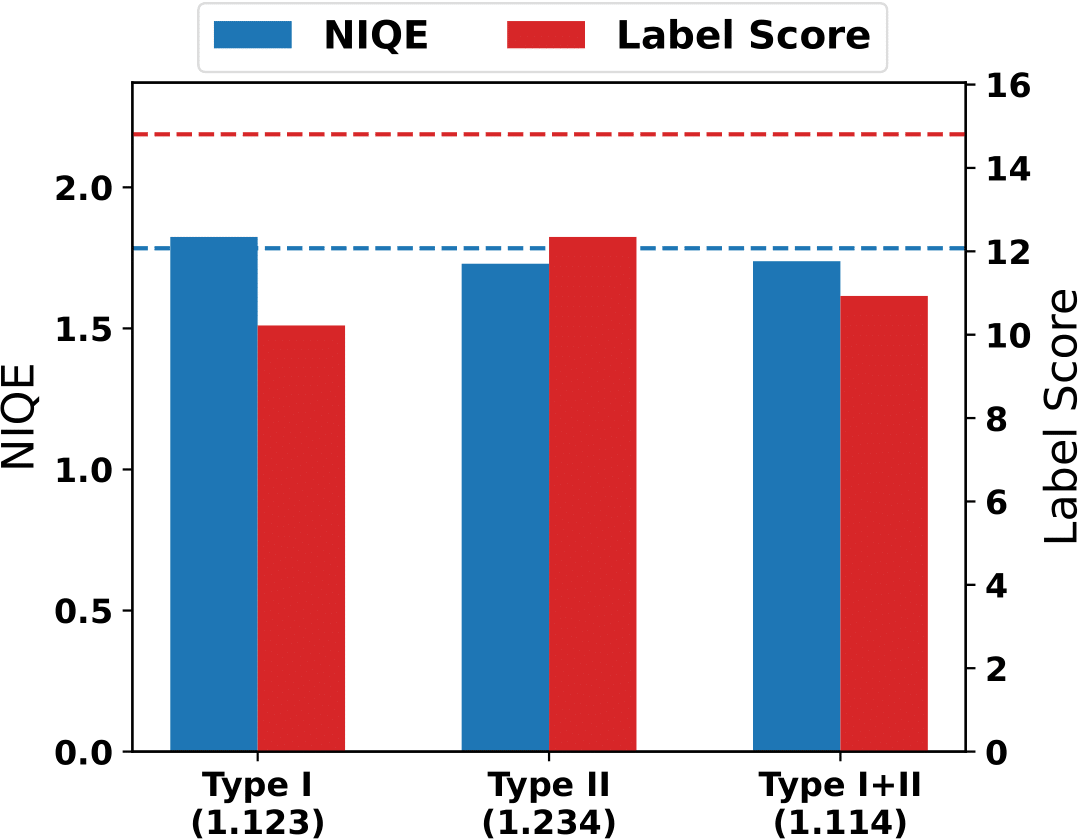}
	\caption{The individual impacts of Type I and Type II negative samples on Dual-NDA in the Steering Angle (64$\times$64) experiment. The SFID scores for compared settings are shown in parentheses. The blue and red dashed lines represent the NIQE and Label Score of ``CcGAN w/o NDA", respectively. }
	\label{fig:SA64_ablation_effect_types}
\end{figure}

\begin{figure}[!h] 
	\centering
	\includegraphics[width=0.8\linewidth]{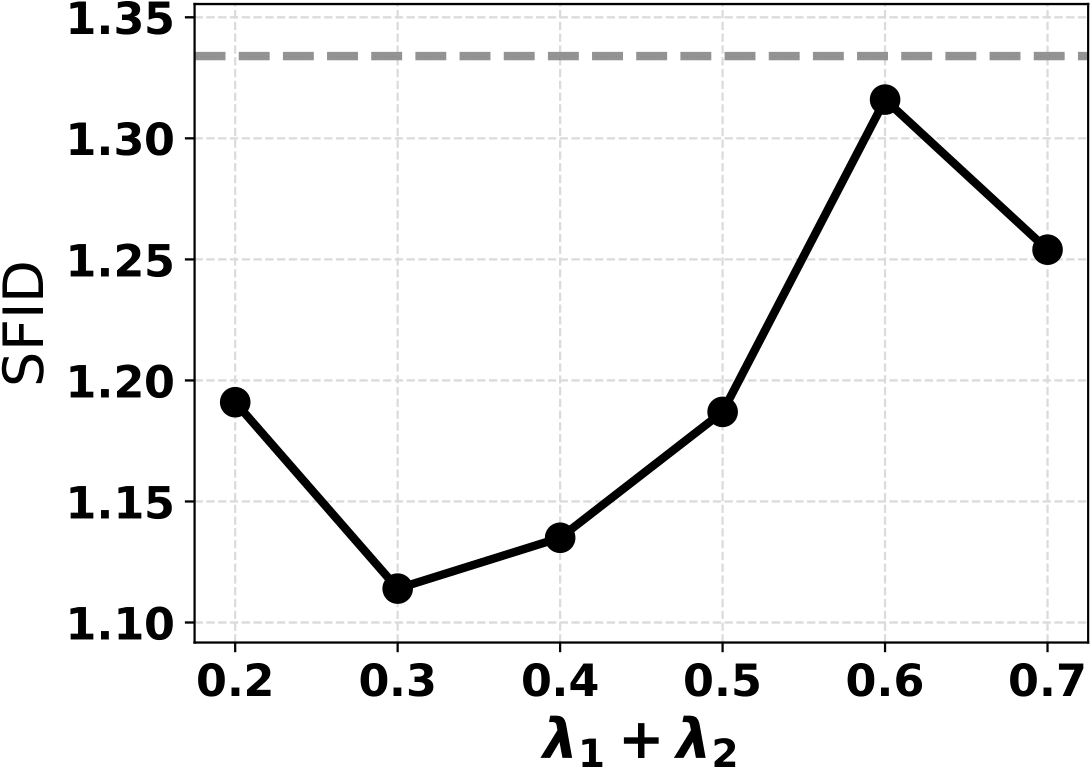}
	\caption{The effect of $\lambda_1+\lambda_2$ on Dual-NDA in the Steering Angle (64$\times$64) experiment. The gray dashed line represents the SFID score of ``CcGAN w/o NDA". }\label{fig:SA64_ablation_effect_lambda}
\end{figure}

\begin{table*}[!h]
	\centering
	\small
	\begin{tabular}{ccccc}
		\toprule
		$\bm{q_1}$    & \textbf{SFID}  & \textbf{NIQE} & \textbf{Diversity} & \textbf{Label Score} \\
		\midrule
		w/o NDA  & 1.334 & 1.784 & 1.234 & 14.807 \\\hline
		0.3   & 1.125 & 1.759 & 1.271 & 11.365 \\
		0.5   & 1.114 & 1.738 & 1.251 & 11.809 \\
		0.7   & 1.161 & 1.763 & 1.227 & 11.162 \\
		0.9   & 1.187 & 1.763 & 1.270 & 11.550 \\
		\bottomrule
	\end{tabular}%
	\caption{The impact of parameter $q_1$ on the Dual-NDA performance in the Steering Angle (64x64) experiment. ``CcGAN w/o NDA" is included in the table as a baseline for reference.}
	\label{tab:effect_of_q1}%
\end{table*}%

\begin{table*}[!htbp]
	\centering
	\small
	\begin{tabular}{ccccc}
		\toprule
		$\bm{q_2}$    & \textbf{SFID} & \textbf{NIQE} & \textbf{Diversity} & \textbf{Label Score} \\
		\midrule
		w/o NDA  & 1.334 & 1.784 & 1.234 & 14.807 \\\hline
		0.5   &  1.107 & 1.717 &  1.301 & 11.881 \\
		0.6   &  1.193 & 1.733 &  1.278 & 11.459 \\
		0.7   &  1.150 & 1.763 &  1.278 & 13.161 \\
		0.8   &  1.174 & 1.756 &  1.254 & 11.207 \\
		0.9   &  1.114 & 1.738 &  1.251 & 11.809 \\
		\bottomrule
	\end{tabular}%
	\caption{The impact of parameter $q_2$ on the Dual-NDA performance in the Steering Angle (64x64) experiment. ``CcGAN w/o NDA" is included in the table as a baseline for reference.}
	\label{tab:effect_of_q2}%
\end{table*}%

\begin{table*}[!h]
	\centering
	\small
	\begin{tabular}{ccccc}
		\toprule
		\textbf{Generator}  & \textbf{SFID} & \textbf{NIQE} & \textbf{Diversity} & \textbf{Label Score} \\
		\midrule
		None   & 1.334 & 1.784 & 1.234 & 14.807 \\\midrule
		ADCGAN   &  1.212 & \textbf{1.832}   & 1.222 & 10.884 \\
		ReACGAN   &  1.203 & \textbf{1.800} & 1.273 & 11.486 \\
		\textbf{CcGAN}  & 1.114 & 1.738 & 1.251 & 11.809 \\
		\bottomrule
	\end{tabular}%
	\caption{Type II negative samples from different generators. }
	\label{tab:different_generators}%
\end{table*}%

\section{Conclusion}

In this paper, we present an innovative NDA approach called Dual-NDA, aimed at enhancing the performance of CcGANs in the realm of continuous conditional generative modeling. Our approach involves two strategies for generating negative samples, simulating two categories of low-quality images that may arise during CcGAN sampling. Furthermore, we introduce a modified CcGAN mechanism that relies on these negative samples to steer the model away from generating undesirable outputs. Through comprehensive experimentation, we demonstrate that Dual-NDA effectively improves the performance of CcGANs, surpassing widely-used class-conditional GANs and diffusion models.

\section{Acknowledgments}
This work was supported by the National Natural Science Foundation of China (Grant No.~62306147) and Pujiang Talent Program (Grant No.~23PJ1412100).

\fontsize{9.5pt}{10.5pt} 
\selectfont
\bibliography{reference}

\clearpage
\newpage
\appendix

\setcounter{secnumdepth}{2}

\section*{Appendix}
%
%
\section{Detailed Algorithms for Dual-NDA}\label{supp:detailed_algorithm}

\subsection{The Two Algorithms for Generating Type II Negative Samples}

We employ two distinct NIQE filtering algorithms, as detailed in \cref{alg:type_2_int} and \ref{alg:type_2_cont}, for generating Type II negative samples. Specifically, \cref{alg:type_2_int} is designed to handle regression labels characterized by integer-valued observations (e.g., ages), whereas \cref{alg:type_2_cont} is crafted to address regression labels characterized by strictly continuous values (e.g., angles). It is essential to emphasize that the process of generating Type II negative samples is carried out offline.

\begin{algorithm}[!h]
	\caption{The NIQE filtering algorithm for generating the Type II negative samples with integer-valued regression labels (e.g., ages). Assume the integer-valued regression label of interest, denoted as $y$, spans from $1$ to $K$.}
	\label{alg:type_2_int}  
	\KwData{Randomly sample $M$ fake image-label pairs from a pre-trained generator, i.e., $Q^g=\{(\bm{x}_i^g,y_i^g)| i=1,...,M\}$;}
	\KwIn{NIQE models pretrained on the training set of CcGANs; The preset hyperparameter $q_2$;}
	\KwResult{$Q_\text{II} = \{ (\tilde{\bm{x}}_i^{g}, \tilde{y}_i^{g})|  \text{NIQE}(\tilde{\bm{x}}_i^{g})>c_{q_2}, (\tilde{\bm{x}}_i^{g}, \tilde{y}_i^{g})\in Q^g, i=1,...,N^g_\text{II}, N^g_\text{II}<M \}$}
	$k\leftarrow 1$, $Q_\text{II} = \emptyset$\;
	\While{$k \leq K$}{
		\tcp{Conduct filtering for each $k$}
		Denote fake images with label $k$ as $\bm{X}^g_{k}=\allowbreak\{\bm{x}_{1}^{g,k},..., \bm{x}_{M_{k}}^{g,k} \}$ \;
		Compute the NIQE scores of images in $\bm{X}_{k}$, and store the scores in an array, denoted as $\bm{s}^k=[s_1^k,s_2^k,...,s_{M_{k}}^k]$\;
		Calculate the $q_2$-th quantile of $\bm{s}^k$, denoted as $c^k_{q_2}$\;
		\For{$i\leftarrow 1$ \KwTo $M_{k}$}{ 
			\tcp{Do the NIQE filtering}
			\If{ $NIQE({\bm{x}}_{i}^{g,k})>c^k_{q_2}$}{
				$Q_\text{II}\leftarrow Q_\text{II}\cup \{ ({\bm{x}}_{i}^{g,k}, k) \}$
			}
		}
		$k\leftarrow k+1$\;}
\end{algorithm}

\begin{algorithm}[!h]
	\caption{The NIQE filtering algorithm for generating the Type II negative samples with strictly continuous regression labels (e.g., angles).}
	\label{alg:type_2_cont}   
	\KwData{Randomly sample $M$ fake image-label pairs from a pre-trained generator, i.e., $Q^g=\{(\bm{x}_i^g,y_i^g)| i=1,...,M\}$;}
	\KwIn{NIQE models pretrained on the training set of CcGANs; The preset hyperparameter $q_2$;}
	\KwResult{$Q_\text{II} = \{ (\tilde{\bm{x}}_i^{g}, \tilde{y}_i^{g})|  \text{NIQE}(\tilde{\bm{x}}_i^{g})>c_{q_2}, (\tilde{\bm{x}}_i^{g}, \tilde{y}_i^{g})\in Q^g, i=1,...,N^g_\text{II}, N^g_\text{II}<M \}$}
	Compute the NIQE scores of images in $Q^g$, and store them in an array, denoted as $\bm{s}=[s_1,s_2,...,s_{M}]$\;
	Calculate the $q_2$-th quantile of $\bm{s}$, denoted as $c_{q_2}$\;
	\tcp{Do the NIQE filtering}
	$Q_\text{II} = \emptyset$\;
	\For{$i\leftarrow 1$ \KwTo $M$}{ 
		\If{ $NIQE({\bm{x}}_{i}^{g})>c_{q_2}$}{
			$Q_\text{II}\leftarrow Q_\text{II}\cup \{ (\bm{x}_i^g, y_i^g) \}$
		}
	}
\end{algorithm}

\subsection{The Dual-NDA-based Training Mechanism for CcGAN (SVDL+ILI)}

In this section, we propose a new training mechanism for CcGANs (SVDL+ILI)~\citep{ding2023ccgan}, described in \cref{alg:ccgan_with_dual_nda}. The Type I negative samples' generation procedures have been integrated into \cref{alg:ccgan_with_dual_nda}. The CcGAN-related hyperparameters $\sigma$, $\nu$, and $\kappa$ are determined by using the rule of thumb provided by \citet{ding2023ccgan}.

\begin{algorithm*}[!h]
	\footnotesize
	\SetAlgoLined
	\KwData{$N^r$ real image-label pairs $\Omega^r=\{(\bm{x}^r_1, y^r_1),\dots,(\bm{x}^r_{N^r}, y^r_{N^r})\}$, $N_{\text{uy}}^r$ ordered distinct labels $\Upsilon=\{y_{[1]}^r, \dots, y_{[N_{\text{uy}}^r]}^r \}$ in the dataset, ${N^g_\text{II}}$ pre-generated Type II fake image-label pairs $Q_\text{II} = \{ (\tilde{\bm{x}}_1^{g}, \tilde{y}_1^{g}), ..., (\tilde{\bm{x}}_{N^g_\text{II}}^{g}, \tilde{y}_{N^g_\text{II}}^{g}) \}$, preset $\sigma$, $\nu$, $\kappa$, $q_1$, $\lambda_1$, and $\lambda_2$, number of iterations $K$, the discriminator batch size $m^d$, and the generator batch size $m^g$.}
	\KwResult{A trained generator $G$ and a trained discriminator $D$.} 
	\For{$k=1$ \KwTo $K$}{
		\tcp{Update $D$'s parameters}
		Draw $m^d$ labels $Y^d$ with replacement from $\Upsilon$\;
		Create a set of target labels $Y^{d,\epsilon}=\{ y_i+\epsilon| y_i\in Y^d, \epsilon\in \mathcal{N}(0,\sigma^2), i=1,\dots,m^d \}$ ($D$ is conditional on these labels)  \;
		Initialize $\Omega_d^r=\phi$, $\Omega_d^f=\phi$, $\Omega_d^\text{I}=\phi$, $\Omega_d^\text{II}=\phi$\;
		\For{$i=1$ \KwTo $m^d$}{
			\tcp{Draw one real image from $\Omega_d^r=\phi$ with label in a \textbf{soft} vicinity of $y_i+\epsilon$.}
			Randomly choose an image-label pair $(\bm{x},y) \in \Omega^r$ satisfying $e^{-\nu (y-y_i-\epsilon)^2}>10^{-3}$ where $y_i+\epsilon\in Y^{d,\epsilon}$ and let $\Omega_d^r=\Omega_d^r \cup (\bm{x}, y_i+\epsilon)$ \;
			Compute $w_i^r(y, y_i + \epsilon)=e^{-\nu (y_i + \epsilon - y)^2 }$\;
			
			\tcp{Generate one fake image from the fixed generator network given $y_i+\epsilon$.}
			Randomly draw a label $y^\prime$ from $U(y_i+\epsilon-\sqrt{-\frac{\log 10^{-3}}{\nu}}, y_i+\epsilon+\sqrt{-\frac{\log 10^{-3}}{\nu}})$\;
			Generate a fake image $\bm{x}^\prime$ by evaluating $G(\bm{z}, y^\prime)$, where $\bm{z}\sim \mathcal{N}(\bm{0},\bm{I})$ and let $\Omega_d^f=\Omega_d^f \cup (\bm{x}^\prime, y_i+\epsilon)$ \;
			Compute $w_i^g(y^\prime, y_i + \epsilon)=e^{-\nu (y_i + \epsilon - y^\prime)^2 }$\;
			
			\tcp{Generate one Type I negative image \textbf{in an online manner}.}
			Compute the pairwise absolute distances between $y_i + \epsilon$ and the labels of all real images in the training set, denoted as $d_j^{y_i + \epsilon}=|y_i + \epsilon-y_j^r|$, $j=1,...,N^r$. These absolute distance values form an array, denoted as $\bm{d}^{y_i+\epsilon}=[d_1^{y_i+\epsilon},...,d_{N^r}^{y_i+\epsilon}]$.\;
			Calculate the $q_1$-th quantile of $\bm{d}^{y_i+\epsilon}$, denoted as $c_{q_1}^{y_i+\epsilon}$\;
			Choose training images with labels greater than $c_{q_1}^{y_i+\epsilon}$ to form a set, denoted as $\bm{Q}_\text{I}^{y_i+\epsilon} = \{(\tilde{\bm{x}}_i^{r}, \tilde{y}_i^{r})| (\tilde{\bm{x}}_i^{r}, \tilde{y}_i^{r})\in\Omega^r, |\tilde{y}_i^{r}-y_i-\epsilon|>c_{q_1}^{y_i+\epsilon},  i=1,...,N^r_{y_i+\epsilon} \}$ \;
			Randomly choose one image-label pair $(\tilde{\bm{x}}^r,\tilde{y}^r)$ from $\bm{Q}_\text{I}^{y_i+\epsilon}$ and let $\Omega_d^\text{I}=\Omega_d^\text{I} \cup (\tilde{\bm{x}}^r, y_i+\epsilon)$ \;
			
			\tcp{Draw one Type II negative image from $Q_\text{II}$ with label in a \textbf{hard} vicinity of $y_i+\epsilon$.}
			Randomly choose an image-label pair $(\tilde{\bm{x}}^g,\tilde{y}^g) \in Q_\text{II}$ satisfying $|\tilde{y}^g-y_i-\epsilon|\leq\kappa$ where $y_i+\epsilon\in Y^{d,\epsilon}$ and let $\Omega_d^\text{II}=\Omega_d^\text{II} \cup (\tilde{\bm{x}}^g, y_i+\epsilon)$. \;
		}
		Update $D$ with samples in set $\Omega_d^r$, $\Omega_d^f$, $\Omega_d^\text{I}$ , and $\Omega_d^\text{II}$ via gradient-based optimizers based on Eq.(7)\; 
		
		\tcp{Update $G$'s parameters}
		Draw $m^g$ labels $Y^g$ with replacement from $\Upsilon$\;
		Create another set of target labels $Y^{g,\epsilon}=\{ y_i+\epsilon| y_i\in Y^g, \epsilon\in \mathcal{N}(0,\sigma^2), i=1,\dots,m^g \}$ ($G$ is conditional on these labels) \;
		Generate $m^g$ fake images conditional on $Y^{g,\epsilon}$ and put these image-label pairs in $\Omega_g^f$ \;
		Update $G$ with samples in $\Omega_g^f$ via gradient-based optimizers based on Eq.(2) \;
		
	}
	\caption{A modified training mechanism for CcGAN (SVDL+ILI) with Dual-NDA.}
	\label{alg:ccgan_with_dual_nda}
\end{algorithm*}

\subsection{Reformulated Hinge Loss}

Following~\citet{ding2023ccgan}, the vicinal discriminator loss introduced in this study (Eq. (7)) is derived in terms of the vanilla cGAN loss~\citep{mirza2014conditional}. Furthermore, in accordance with Eq. (S.26) in the Appendix of \citet{ding2023ccgan}, we present an alternative formulation of Eq. (7) using the hinge loss, as demonstrated below:
\begin{equation}
	\label{eq:new_vicinal_disc_loss_hinge}
	\scalemath{0.95}{
		\begin{aligned}
			& \widetilde{\mathcal{L}}(D) = \\
			-& \frac{1}{N^r}\sum_{j=1}^{N^r}\sum_{i=1}^{N^r}\mathbb{E}_{\epsilon\sim\mathcal{N}(0,\sigma^2)}\left[W_1\cdot\min( 0, -1+D(\bm{x}_i^r, y_j^r+\epsilon) ) \right] \\
			- & \frac{1-\bar{\lambda}}{N^g}\sum_{j=1}^{N^g}\sum_{i=1}^{N^g}\mathbb{E}_{\epsilon\sim\mathcal{N}(0,\sigma^2)}\left[ W_2\cdot \min( 0,-1-D(\bm{x}_i^g, y_j^g+\epsilon) ) \right], \\
			-&\frac{\lambda_1}{N_\text{I}}\sum_{i=1}^{N_\text{I}}\min(0,-1-D(\tilde{\bm{x}}_i^{r}, y_i^\text{I}))) \\
			-&\frac{\lambda_2}{N_\text{II}}\sum_{j=1}^{N_\text{II}}\sum_{i=1}^{N_\text{II}}\mathds{E}_{\epsilon\sim\mathcal{N}(0,\sigma^2)}\left[ W_3\cdot\min\left(0,-1-D\left(\tilde{\bm{x}}_i^{g}, \tilde{y}_j^{g}+\epsilon\right)\right) \right].
		\end{aligned}
	}	
\end{equation}
The corresponding generator loss is also provided as follows:
\begin{equation}
	\label{eq:gen_loss_hinge}
	\scalemath{0.95}{
		\widehat{\mathcal{L}}(G) = - \frac{1}{N^g}\sum_{i=1}^{N^g}\mathds{E}_{\epsilon\sim\mathcal{N}(0,\sigma^2)} D(G(\bm{z}_{i}, y_i^g+\epsilon), y_i^g+\epsilon).
	}
\end{equation}
It is worth noting that \cref{alg:ccgan_with_dual_nda} can be appropriately adapted if the above hinge loss is utilized.

\section{Detailed Experimental Setups}\label{supp:detailed_setups}

\subsection{GitHub Repository}
The codebase for this work will be released at:
\begin{center}
	\url{https://github.com/UBCDingXin/Dual-NDA}
\end{center}

\subsection{The Data Distribution of UTKFace and Steering Angle}

All samples from both the UTKFace and Steering Angle datasets are employed for training purposes. The distribution of these samples is depicted in \Cref{fig:data_distribution}, revealing pronounced data imbalance issues within both datasets. This issue becomes more severe in the Steering Angle datasets, where some steering angle values are only represented by less than 10 images. 

\begin{figure*}[!h] 
	\centering
	\includegraphics[width=0.7\linewidth]{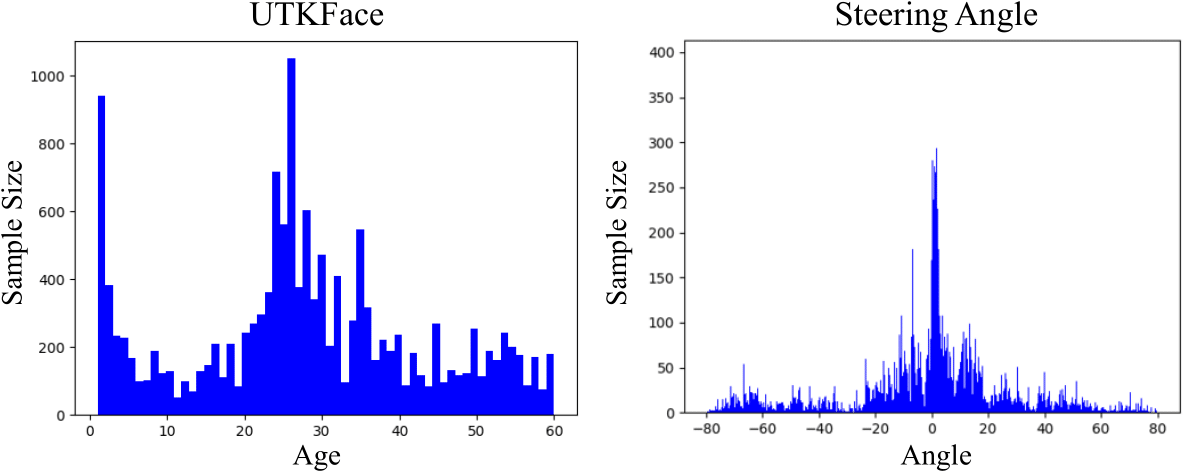}
	\caption{The data distribution of the UTKFace and Steering Angle datasets.}\label{fig:data_distribution}
\end{figure*}

\subsection{Detailed Training Setups}

The baseline methods include ReACGAN~\citep{kang2021rebooting}, ADCGAN~\citep{hou2022conditional}, ADM-G~\citep{dhariwal2021diffusion}, CFG~\citep{ho2021classifier}, CcGAN (SVDL+ILI)~\citep{ding2023ccgan}, CcGAN (SVDL+ILI) w/ NDA~\citep{sinha2021negative}, and the proposed Dual-NDA. We leverage the following code repositories to implement these baseline methods:
\begin{itemize}
	\item \textbf{ReACGAN and ADCGAN}: \\
	\url{https://github.com/POSTECH-CVLab/PyTorch-StudioGAN}
	
	\item \textbf{ADM-G}:\\
	\url{https://github.com/openai/guided-diffusion}
	
	\item \textbf{CFG}:\\
	\url{https://github.com/lucidrains/denoising-diffusion-pytorch}
	
	\item \textbf{CcGAN (SVDL+ILI)}:\\
	\url{https://github.com/UBCDingXin/improved_CcGAN}
	
	\item \textbf{CcGAN (SVDL+ILI) w/ NDA}:\\
	\url{https://github.com/ermongroup/NDA}
	
\end{itemize}

We have released our codebase, accompanied by a comprehensive \texttt{README.md} file. The \texttt{README.md} file contains detailed descriptions of our most important training setups. These setups are outlined in both the \texttt{README.md} file and the training batch scripts (\texttt{*.bat} or \texttt{*.sh} files) within our code repository. Here, we list some important setups in \Cref{tab:ccgan_setup}. 

\begin{table*}[!h] 
	\centering
	\begin{adjustbox}{width=1\textwidth}
		\begin{tabular}{ccl}
			\toprule
			\textbf{Dataset} & \textbf{Method} & \textbf{Setup} \\
			\midrule
			
			\multirow{7}[16]{*}{\begin{tabular}[c]{@{}c@{}} \textbf{UTKFace} \\ {(64$\times$64)}\end{tabular}} & ReACGAN (Neurips'21) & \begin{tabular}[c]{@{}l@{}} big resnet, hinge loss, steps=$40K$, batch size=256, update $D$ twice per step, $\text{lr}_D=2\times 10^{-4}$, $\text{lr}_G=5\times 10^{-5}$\end{tabular}   \\\cline{2-3}
			& ADCGAN (ICML'22) & \begin{tabular}[c]{@{}l@{}} big resnet, hinge loss, steps=$20K$, batch size=128, update $D$ twice per step, $\text{lr}_D=2\times 10^{-4}$, $\text{lr}_G=5\times 10^{-5}$\end{tabular} \\\cline{2-3}
			& ADM-G (Neurips'21) & \begin{tabular}[c]{@{}l@{}} Classifier: steps=$20K$, batch size=128, lr=$3\times 10^{-4}$ \\ Diffusion: steps=65K, batch size=64, lr=$1\times 10^{-5}$, diffusion steps=$1K$  \end{tabular}   \\\cline{2-3}
			& CFG (Neurips'21) & steps=$100K$, lr=$10^{-4}$, batch size=1024, time steps (train/sampling)=1000/100 \\ \cline{2-3}
			
			&  w/o NDA (T-PAMI'23) & \begin{tabular}[c]{@{}l@{}} SNGAN, the vanilla loss, SVDL+ILI, $\sigma=0.041$, $\nu=3600$, use DiffAugment, steps=$40K$, batch size=256, lr=$10^{-4}$, \\ update $D$ twice per step\end{tabular} \\\cline{2-3}
			& w/ NDA (ICLR'21) & \begin{tabular}[c]{@{}l@{}} SNGAN, the vanilla loss, SVDL+ILI, $\sigma=0.041$, $\nu=3600$,  use DiffAugment, steps=$40K$, batch size=256, lr=$10^{-4}$, \\ update $D$ twice per step, use JigSaw to create negative samples \end{tabular} \\\cline{2-3}
			& Dual-NDA & \begin{tabular}[c]{@{}l@{}} SNGAN, the vanilla loss, SVDL+ILI, $\sigma=0.041$, $\nu=3600$, $\kappa=0.017$, use DiffAugment, steps=$60K$, Dual-NDA starts at $40K$,  \\ batch size=256, lr=$10^{-4}$, update $D$ twice per step, 
				$\lambda_1=0.05$, $q_1=0.9$, $\lambda_2=0.15$, $q_2=0.9$, $N_\text{II}=60K$ ($1K$ per age value)\end{tabular} \\ \midrule
			
			\multirow{7}[16]{*}{\begin{tabular}[c]{@{}c@{}} \textbf{UTKFace} \\ {(128$\times$128)}\end{tabular}} & ReACGAN (Neurips'21) & \begin{tabular}[c]{@{}l@{}} big resnet, hinge loss, steps=$20K$, batch size=128, update $D$ twice per step, $\text{lr}_D=2\times 10^{-4}$, $\text{lr}_G=5\times 10^{-5}$\end{tabular} \\\cline{2-3}
			& ADCGAN (ICML'22) & \begin{tabular}[c]{@{}l@{}} big resnet, hinge loss, steps=$20K$, batch size=128, update $D$ twice per step, $\text{lr}_D=2\times 10^{-4}$, $\text{lr}_G=5\times 10^{-5}$, use DiffAugment \end{tabular} \\ \cline{2-3}
			& ADM-G (Neurips'21) &  \begin{tabular}[c]{@{}l@{}} Classifier: steps=$20K$, batch size=64, lr=$3\times 10^{-4}$ \\ Diffusion: steps=50K, batch size=24, lr=$1\times 10^{-5}$, diffusion steps=$1K$  \end{tabular} \\\cline{2-3}
			& CFG (Neurips'21) & steps=$50K$, lr=$10^{-5}$, batch size=64, time steps (train/sampling)=1000/100 \\ \cline{2-3}
			
			& w/o NDA (T-PAMI'23) & \begin{tabular}[c]{@{}l@{}} SAGAN, the hinge loss, SVDL+ILI, $\sigma=0.041$, $\nu=900$, use DiffAugment, steps=$20K$, batch size=256, lr=$10^{-4}$, \\ update $D$ four times per step\end{tabular} \\\cline{2-3}
			& w/ NDA (ICLR'21) & \begin{tabular}[c]{@{}l@{}} SAGAN, the hinge loss, SVDL+ILI, $\sigma=0.041$, $\nu=900$, use DiffAugment, steps=$20K$, batch size=256, lr=$10^{-4}$, \\ update $D$ four times per step, use JigSaw to create negative samples \end{tabular} \\\cline{2-3}
			& Dual-NDA & \begin{tabular}[c]{@{}l@{}} SAGAN, the hinge loss, SVDL+ILI, $\sigma=0.041$, $\nu=900$, $\kappa=0.033$, use DiffAugment, steps=$22500$, Dual-NDA starts at $20K$, \\ batch size=256, lr=$10^{-4}$, update $D$ four times per step, 
				$\lambda_1=0.05$, $q_1=0.9$, $\lambda_2=0.15$, $q_2=0.9$, $N_\text{II}=60K$ ($1K$ per age value)\end{tabular} \\ \midrule

			\multirow{7}[20]{*}{\begin{tabular}[c]{@{}c@{}} \textbf{Steering Angle} \\ {(64$\times$64)}\end{tabular}} & ReACGAN (Neurips'21) & \begin{tabular}[c]{@{}l@{}} big resnet, hinge loss, steps=$20K$, batch size=256, update $D$ twice per step, $\text{lr}_D=2\times 10^{-4}$, $\text{lr}_G=5\times 10^{-5}$\end{tabular}   \\\cline{2-3}
			& ADCGAN (ICML'22) & \begin{tabular}[c]{@{}l@{}} big resnet, hinge loss, steps=$20K$, batch size=128, update $D$ twice per step, $\text{lr}_D=2\times 10^{-4}$, $\text{lr}_G=5\times 10^{-5}$\end{tabular} \\\cline{2-3}
			& ADM-G (Neurips'21) & \begin{tabular}[c]{@{}l@{}} Classifier: steps=$20K$, batch size=128, lr=$3\times 10^{-4}$ \\ Diffusion: steps=50K, batch size=32, lr=$3\times 10^{-4}$, diffusion steps=$4K$  \end{tabular}   \\\cline{2-3}
			& CFG (Neurips'21) & steps=$80K$, lr=$10^{-4}$, batch size=128, time steps (train/sampling)=1000/100 \\ \cline{2-3}
			
			&  w/o NDA (T-PAMI'23) & \begin{tabular}[c]{@{}l@{}} SAGAN, the hinge loss, SVDL+ILI, $\sigma=0.029$, $\nu=1000.438$, use DiffAugment, steps=$20K$, batch size=512, lr=$10^{-4}$, \\ update $D$ twice per step\end{tabular} \\\cline{2-3}
			& w/ NDA (ICLR'21) & \begin{tabular}[c]{@{}l@{}} SAGAN, the hinge loss, SVDL+ILI, $\sigma=0.029$, $\nu=1000.438$, use DiffAugment, steps=$20K$, batch size=512, lr=$10^{-4}$, \\ update $D$ twice per step, use JigSaw to create negative samples \end{tabular} \\\cline{2-3}
			& Dual-NDA & \begin{tabular}[c]{@{}l@{}} SAGAN, the hinge loss, SVDL+ILI, $\sigma=0.029$, $\nu=1000.438$, $\kappa=0.032$, use DiffAugment, steps=$20K$, Dual-NDA starts at 0, \\ batch size=512, lr=$10^{-4}$, update $D$ twice per step, 
				$\lambda_1=0.1$, $q_1=0.5$, $\lambda_2=0.2$, $q_2=0.9$, \\ $N_\text{II}=17740$ ($10$ Type II negative images for 1774 distinct training angle values)\end{tabular} \\ \midrule
			
			\multirow{7}[20]{*}{\begin{tabular}[c]{@{}c@{}} \textbf{Steering Angle} \\ {(128$\times$128)}\end{tabular}} & ReACGAN (Neurips'21) & \begin{tabular}[c]{@{}l@{}} big resnet, hinge loss, steps=$20K$, batch size=128, update $D$ twice per step, $\text{lr}_D=2\times 10^{-4}$, $\text{lr}_G=5\times 10^{-5}$\end{tabular}   \\\cline{2-3}
			& ADCGAN (ICML'22) & \begin{tabular}[c]{@{}l@{}} big resnet, hinge loss, steps=$20K$, batch size=128, update $D$ twice per step, $\text{lr}_D=2\times 10^{-4}$, $\text{lr}_G=5\times 10^{-5}$, use DiffAugment \end{tabular} \\\cline{2-3}
			& ADM-G (Neurips'21) & \begin{tabular}[c]{@{}l@{}} Classifier: steps=$20K$, batch size=64, lr=$3\times 10^{-4}$ \\ Diffusion: steps=50K, batch size=24, lr=$1\times 10^{-5}$, diffusion steps=$1K$  \end{tabular}   \\\cline{2-3}
			& CFG (Neurips'21) & steps=$50K$, lr=$10^{-5}$, batch size=64, time steps (train/sampling)=1000/100 \\ \cline{2-3}
			
			&  w/o NDA (T-PAMI'23) & \begin{tabular}[c]{@{}l@{}} SAGAN, the hinge loss, SVDL+ILI, $\sigma=0.029$, $\nu=1000.438$, use DiffAugment, steps=$20K$, batch size=256, lr=$10^{-4}$, \\ update $D$ twice per step\end{tabular} \\\cline{2-3}
			& w/ NDA (ICLR'21) & \begin{tabular}[c]{@{}l@{}} SAGAN, the hinge loss, SVDL+ILI, $\sigma=0.029$, $\nu=1000.438$, use DiffAugment, steps=$20K$, batch size=256, lr=$10^{-4}$, \\ update $D$ twice per step, use JigSaw to create negative samples \end{tabular} \\\cline{2-3}
			& Dual-NDA & \begin{tabular}[c]{@{}l@{}} SAGAN, the hinge loss, SVDL+ILI, $\sigma=0.029$, $\nu=1000.438$, $\kappa=0.032$, use DiffAugment, steps=$20K$, Dual-NDA starts at $15K$, \\ batch size=256, lr=$10^{-4}$, update $D$ twice per step, 
				$\lambda_1=0.2$, $q_1=0.5$, $\lambda_2=0.3$, $q_2=0.9$, \\ $N_\text{II}=17740$ ($10$ Type II negative images for 1774 distinct training angle values)\end{tabular} \\
			\bottomrule
		\end{tabular}%
	\end{adjustbox}
	\caption{ Important training setups for baseline methods in Table 1. Please note that, for Dual-NDA, we utilize the pre-trained generator of ``CcGAN w/o NDA" in conjunction with the NIQE filtering process to generate Type II negative samples.}
	\label{tab:ccgan_setup}%
\end{table*}%

\subsection{Detailed Evaluation Setups}

The evaluation setups are almost identical to those adopted by \citep{ding2023ccgan}.

\begin{itemize}
	\item \textbf{UTKFace} \\
	For both $64\times 64$ and $128\times 128$ experiments, we let each compared method generate 1000 fake images for each of 60 age values. Consequently, a total of 60,000 fake image-label pairs are generated for the purpose of evaluation.  We compute the SFID, NIQE, Diversity, and Label Score of these fake samples. The radius for computing SFID is set to zero, so SFID actually decays to Intra-FID \citep{miyato2018cgans} in this case.
	
	\item \textbf{Steering Angle} \\
	The range of steering angles spans from $-80.0^\circ$ and $80.0^\circ$. Within this interval, we generate 2000 evenly spaced numbers as the evaluation angles. For both $64\times 64$ and $128\times 128$ experiments, we let each compared method generate 50 fake images for each of the 2000 evaluation angles. Consequently, a total of 100,000 fake image-label pairs are generated for the purpose of evaluation.  We compute the SFID, NIQE, Diversity, and Label Score of these fake samples. The radius for computing SFID is set to $2^\circ$.

\end{itemize}

\section{Extra Experimental Results}\label{supp:extra_results}

\subsection{The complete results for Tables 2, 3, and 4 in the paper.}

In the paper, owing to space constraints, we do not display the standard deviations for the evaluation scores in Tables 2, 3, and 4. Here, we present the comprehensive results, which are fully detailed in \Cref{tab:effect_of_q1_full}, \Cref{tab:effect_of_q2_full}, and \Cref{tab:different_generators_full}, respectively.

\begin{table*}[!htbp]
	\centering
	\small
	\begin{tabular}{ccccc}
		\toprule
		$\bm{q_1}$    & \textbf{SFID}  & \textbf{NIQE}  & \textbf{Diversity} & \textbf{Label Score} \\
		\midrule
		w/o NDA  & 1.334 (0.531) & 1.784 (0.065) & 1.234 (0.209) & 14.807 (14.297) \\\hline
		0.3   & 1.125 (0.510) & 1.759 (0.047) & 1.271 (0.181) & 11.365 (11.583) \\
		0.5   & 1.114 (0.503) & 1.738 (0.055) & 1.251 (0.172) & 11.809 (11.694) \\
		0.7   & 1.161 (0.540) & 1.763 (0.043) & 1.227 (0.179) & 11.162 (11.241) \\
		0.9   & 1.187 (0.558) & 1.763 (0.045) & 1.270 (0.169) & 11.550 (11.163) \\
		\bottomrule
	\end{tabular}%
	\caption{The impact of parameter $q_1$ on the Dual-NDA performance in the Steering Angle (64x64) experiment. ``CcGAN w/o NDA" is included in the table as a baseline for reference.}
	\label{tab:effect_of_q1_full}%
\end{table*}%

\begin{table*}[!htbp]
	\centering
	\small
	\begin{tabular}{ccccc}
		\toprule
		$\bm{q_2}$    & \textbf{SFID}  & \textbf{NIQE}  & \textbf{Diversity} & \textbf{Label Score} \\
		\midrule
		w/o NDA  & 1.334 (0.531)  & 1.784 (0.065) & 1.234 (0.209)   & 14.807 (14.297) \\\hline
		0.5   &  1.107 (0.506) & 1.717 (0.058) &  1.301 (0.170)  & 11.881 (11.144) \\
		0.6   &  1.193 (0.592) & 1.733 (0.045) &  1.278 (0.205)  & 11.459 (11.675) \\
		0.7   &  1.150 (0.494) & 1.763 (0.038) &  1.278 (0.158)  & 13.161 (13.346) \\
		0.8   &  1.174 (0.563) & 1.756 (0.040) &  1.254 (0.186)  & 11.207 (11.231) \\
		0.9   &  1.114 (0.503) & 1.738 (0.055) & 1.251 (0.172) & 11.809 (11.694) \\
		\bottomrule
	\end{tabular}%
	\caption{The impact of parameter $q_2$ on the Dual-NDA performance in the Steering Angle (64x64) experiment. ``CcGAN w/o NDA" is included in the table as a baseline for reference.}
	\label{tab:effect_of_q2_full}%
\end{table*}%

\begin{table*}[!h]
	\centering
	\small
	\begin{tabular}{ccccc}
		\toprule
		\textbf{Generator}  & \textbf{SFID} $\downarrow$ & \textbf{NIQE} $\downarrow$ & \textbf{Diversity} $\uparrow$ & \textbf{Label Score} $\downarrow$ \\\hline
		None   & 1.334 (0.531) & 1.784 (0.065) & 1.234 (0.209) & 14.807 (14.297) \\\hline
		ADCGAN   &  1.212 (0.481) & \textbf{1.832 (0.055)}   & 1.222 (0.180) & 10.884 (11.041) \\
		ReACGAN   &  1.203 (0.499) & \textbf{1.800 (0.078)} & 1.273 (0.171) & 11.486 (10.904) \\
		\textbf{CcGAN}  & 1.114 (0.503) & 1.738 (0.055) & 1.251 (0.172) & 11.809 (11.694) \\
		\bottomrule
	\end{tabular}%
	\caption{Type II negative samples from different generators. }
	\label{tab:different_generators_full}%
\end{table*}%

\subsection{Some extra ablation studies}
While we recommend using the pre-trained generator of ``CcGAN w/o NDA" to create Type II negative samples, we also conducted experiments using generators from ADCGAN and ReACGAN. ADM-G and CFG were excluded due to their significantly longer sampling times (refer to Fig.~A8 in the Appendix). As shown in Table~\ref{fig:sample_time_SA64}, both ADCGAN and ReACGAN have a detrimental impact on the NIQE values (visual quality) of CcGAN.

The performance of applying Type I and II negative samples separately was originally visualized in Fig.~7 in the paper and is also detailed in \Cref{tab:extra_ablation}. Based on our results in Table 1 and \Cref{tab:extra_ablation}, it appears that NDA tends to degrade the NIQE values of CcGAN. Therefore, we do not recommend combining it with Dual-NDA, and \Cref{tab:extra_ablation} strengthens our recommendation. 

\begin{table*}[!ht]
	\centering
	\small
		\begin{tabular}{ccccc}
			\toprule
			\multicolumn{5}{c}{\textbf{Ablation Study 1}: The effect of various generators for generating Type II } \\
			\multicolumn{5}{c}{ negative samples when implementing Dual-NDA.}\\\hline
			\textbf{Generator}  & \textbf{SFID} $\downarrow$ & \textbf{NIQE} $\downarrow$ & \textbf{Diversity} $\uparrow$ & \textbf{Label Score} $\downarrow$ \\\hline
			None   & 1.334 (0.531) & 1.784 (0.065) & 1.234 (0.209) & 14.807 (14.297) \\\hline
			ADCGAN   &  1.212 (0.481) & \cellcolor{gray!25}{{\color{red} 1.832 (0.055)}}   & 1.222 (0.180) & 10.884 (11.041) \\
			ReACGAN   &  1.203 (0.499) & \cellcolor{gray!25}{{\color{red} 1.800 (0.078)}} & 1.273 (0.171) & 11.486 (10.904) \\
			\textbf{CcGAN (ours)}  & 1.114 (0.503) & 1.738 (0.055) & 1.251 (0.172) & 11.809 (11.694) \\
			\hline\hline
			\multicolumn{5}{c}{\textbf{Ablation Study 2}: Comparison of different combinations of NDA, Type I, and Type II} \\
			\multicolumn{5}{c}{ negative samples. The best result for each metric is marked in boldface. }\\
			\hline
			\textbf{Method}    & \textbf{SFID} $\downarrow$ & \textbf{NIQE} $\downarrow$ & \textbf{Diversity} $\uparrow$ & \textbf{Label Score} $\downarrow$ \\
			\hline
			CcGAN (T-PAMI'23)   & 1.334 (0.531) & 1.784 (0.065) & 1.234 (0.209) & 14.807 (14.297) \\\hline
			w/~NDA (ICLR'21)  & 1.381 (0.527) & \cellcolor{gray!25}{{\color{red} 1.994 (0.081)}}  & 1.231 (0.167) & 10.717 (10.371) \\
			w/~NDA+Type I   &  1.204 (0.564) & 1.892 (0.063) & 1.208 (0.179) & \textbf{9.722 (9.897)} \\
			w/~NDA+Type II   &  1.246 (0.545) & 1.741 (0.045) & 1.284 (0.198) & 10.728 (9.789) \\
			w/~NDA+Type I \& II   & 1.318 (0.570) & 1.746 (0.058)  & 1.191 (0.226) & 9.744 (9.227) \\ \hline
			w/~Type I   &  1.123 (0.478) & 1.824 (0.045)  & 1.228 (0.206) & 10.221 (10.301) \\
			w/~Type II   &  1.234 (0.598) & \textbf{1.729 (0.074)} & \textbf{1.291 (0.186)} & 12.344 (12.153) \\
			\textbf{w/~Type I \& II (ours)} & \textbf{1.114 (0.503)} & 1.738 (0.055)  & 1.251 (0.172) & 11.809 (11.694) \\
			\bottomrule
		\end{tabular}%
	\caption{Two extra ablation studies. }
	\label{tab:extra_ablation}%
\end{table*}%

\subsection{Example Type II Negative Samples}

In this section, we present a selection of example Type II negative samples generated for both the UTKFace (128$\times$128) and Steering Angle (128$\times$128) experiments. The visualization of these samples ranges from \Cref{fig:UK128_Type2_fake_Age1} to \Cref{fig:SA128_Type2_fake_Angle80}. These visuals provide a clear illustration of the generated negative images. It is worth noting that a significant portion of these negative images exhibit evident distortions, artifacts, or are affected by overexposure issues. 

\begin{figure*}[!h] 
	\centering
	\includegraphics[width=1\linewidth]{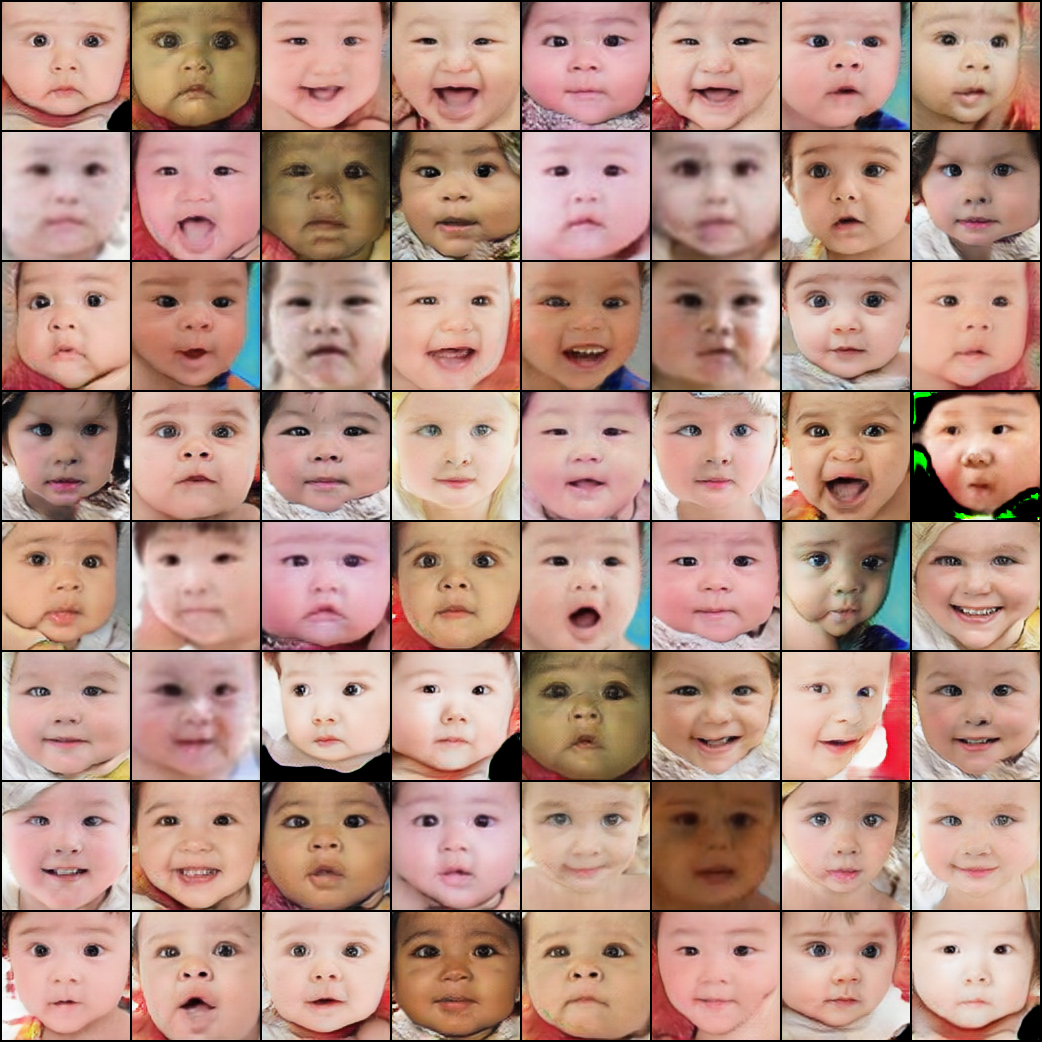}
	\caption{Some example Type II negative images for ``Age=1" in the UTKFace (128$\times$128) experiment.}\label{fig:UK128_Type2_fake_Age1}
\end{figure*}

\begin{figure*}[!h] 
	\centering
	\includegraphics[width=1\linewidth]{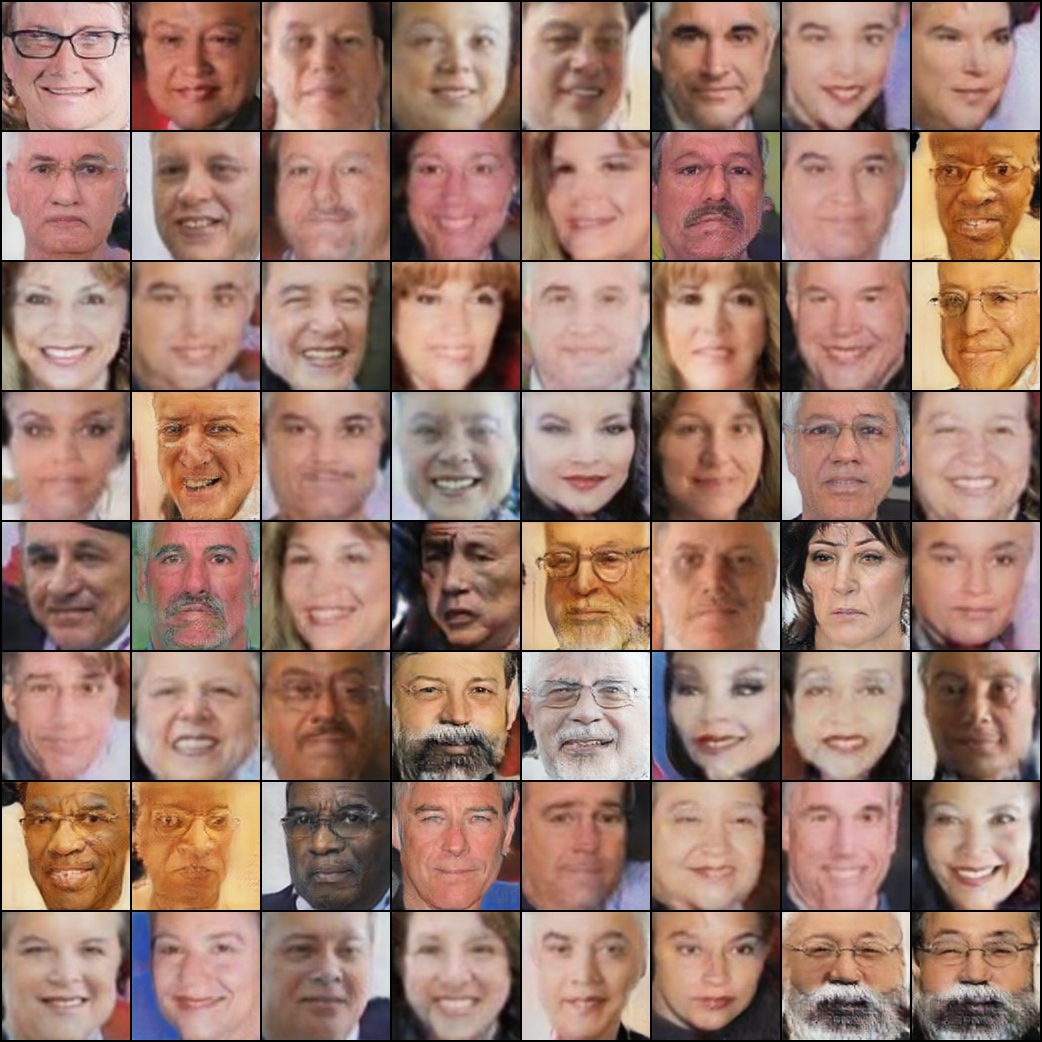}
	\caption{Some example Type II negative images for ``Age=60" in the UTKFace (128$\times$128) experiment.}\label{fig:UK128_Type2_fake_Age60}
\end{figure*}

\begin{figure*}[!h] 
	\centering
	\includegraphics[width=1\linewidth]{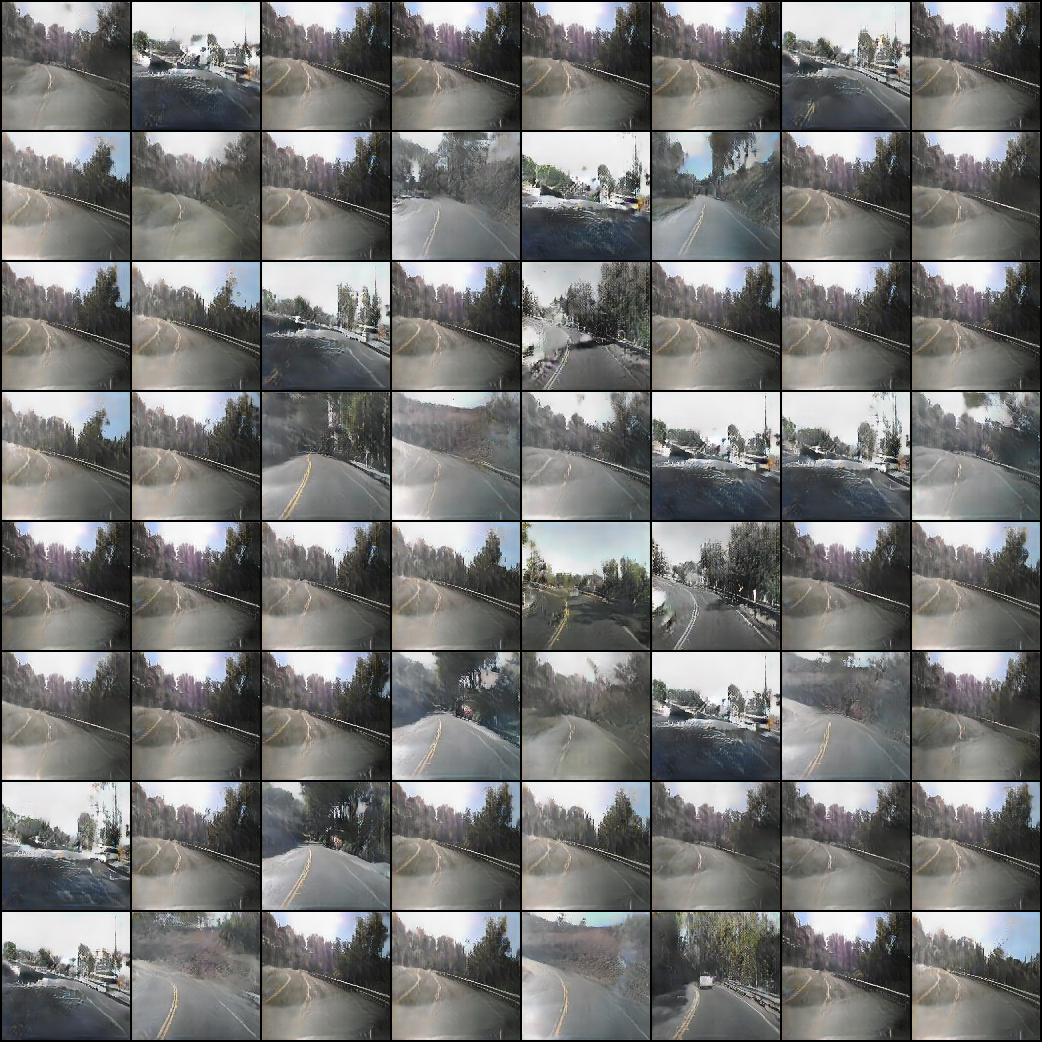}
	\caption{Some example Type II negative images with labels near $-40^\circ$ in the Steering Angle (128$\times$128) experiment.}\label{fig:SA128_Type2_fake_Angle-40}
\end{figure*}

\begin{figure*}[!h] 
	\centering
	\includegraphics[width=1\linewidth]{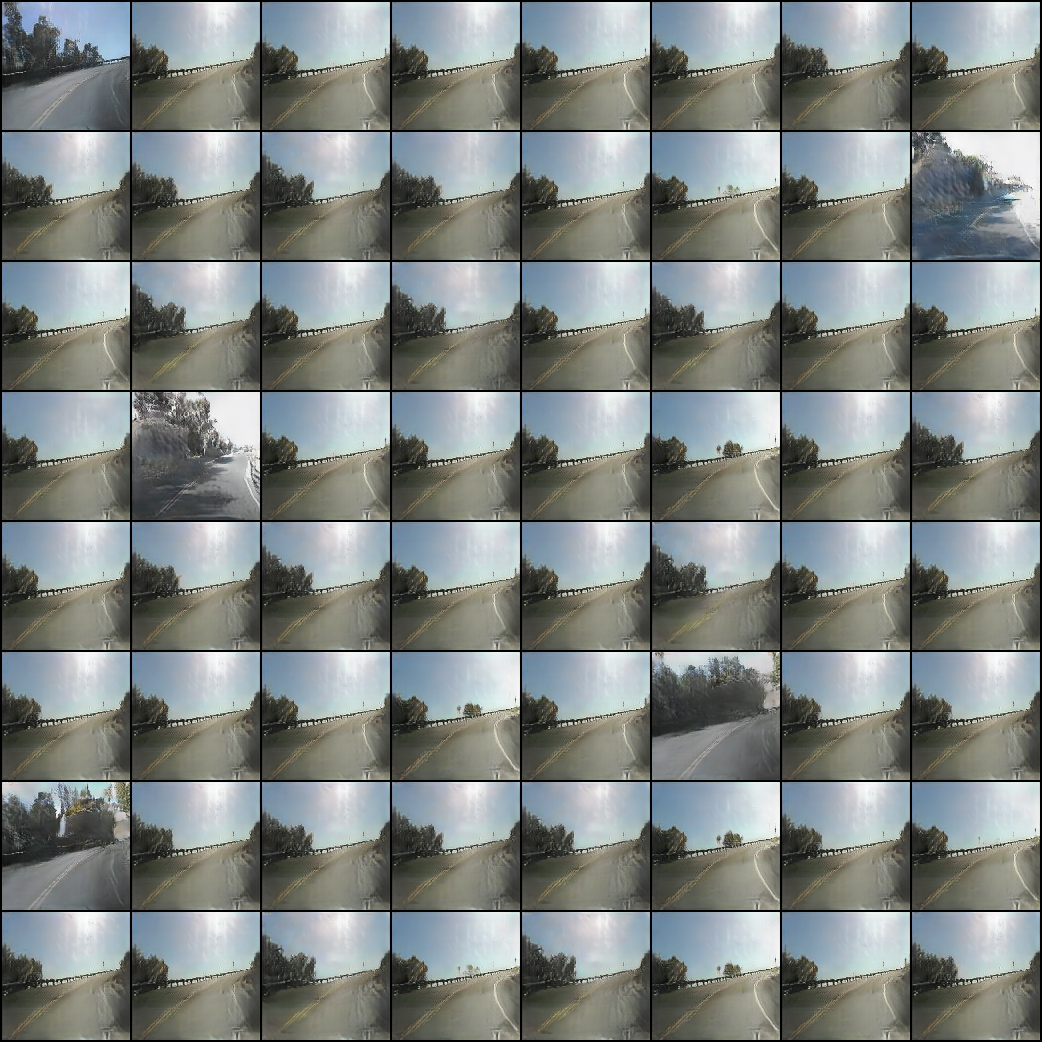}
	\caption{Some example Type II negative images with labels near $80^\circ$ in the Steering Angle (128$\times$128) experiment.}\label{fig:SA128_Type2_fake_Angle80}
\end{figure*}

\subsection{Example Fake Images from Baseline Methods}

In this section, we showcase a selection of example synthetic images generated by both the baseline CcGAN and the proposed Dual-NDA methods. These examples are visualized in \Cref{fig:UK128_example_fake_images} for the UTKFace dataset (128$\times$128) and \Cref{fig:SA128_example_fake_images} for the Steering Angle dataset (128$\times$128). 

For each experimental setting, we intentionally chose three distinct label values that posed challenges for the baseline CcGAN (SVDL+ILI) method. The example images show the baseline CcGAN (SVDL+ILI) often generates fake images marked by subpar visual quality or possibly incorrect labels. However, in contrast, our proposed Dual-NDA consistently excels in producing high-quality synthetic images.

\begin{figure*}[!h] 
	\centering
	\includegraphics[width=0.8\linewidth]{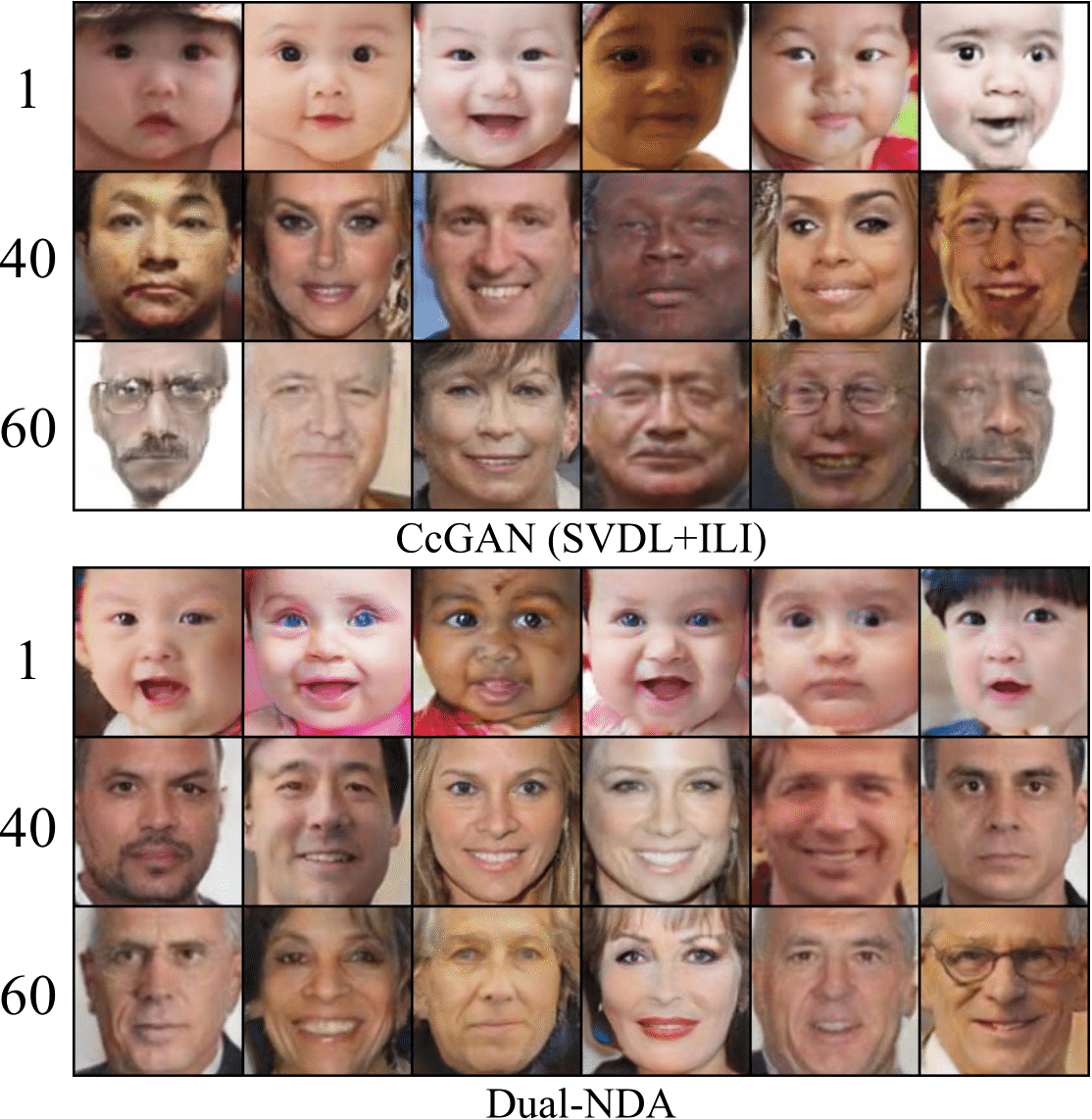}
	\caption{Some example fake images from the baseline CcGAN and Dual-NDA in the UTKFace (128$\times$128) experiment.}\label{fig:UK128_example_fake_images}
\end{figure*}

\begin{figure*}[!h] 
	\centering
	\includegraphics[width=0.8\linewidth]{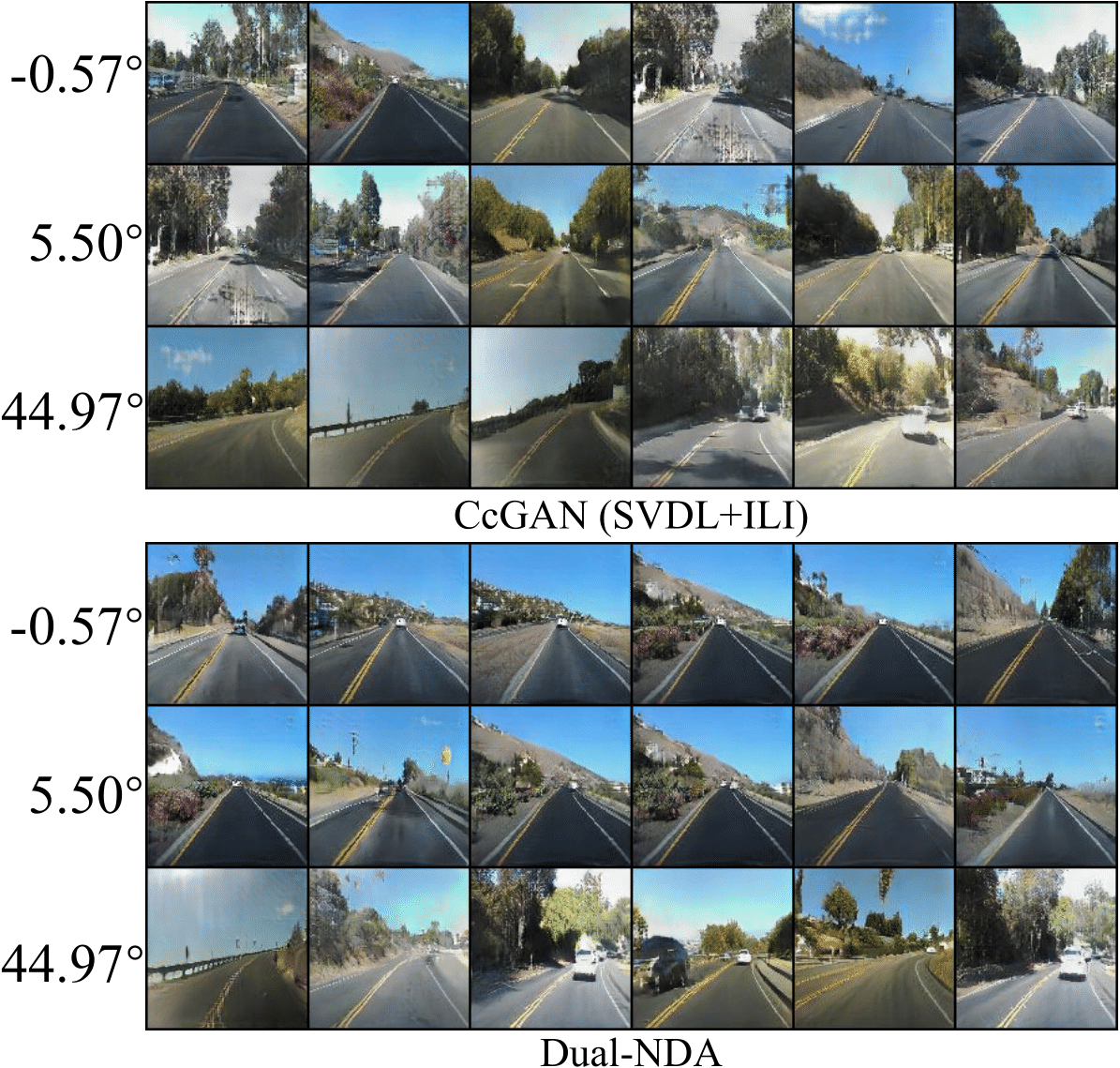}
	\caption{Some example fake images from the baseline CcGAN and Dual-NDA in the Steering Angle (128$\times$128) experiment.}\label{fig:SA128_example_fake_images}
\end{figure*}

\subsection{Sampling Time Comparison}

We also conduct a comparison of the sampling times for the baseline methods on the Steering Angle (64$\times$64) dataset, as depicted in \Cref{fig:sample_time_SA64}. The visualization in the figure reveals that GAN-based methods require only a few seconds to generate 100,000 synthetic images, whereas diffusion models necessitate approximately 6 to 8 hours for the same task.

\begin{figure*}[!h] 
	\centering
	\includegraphics[width=0.7\linewidth]{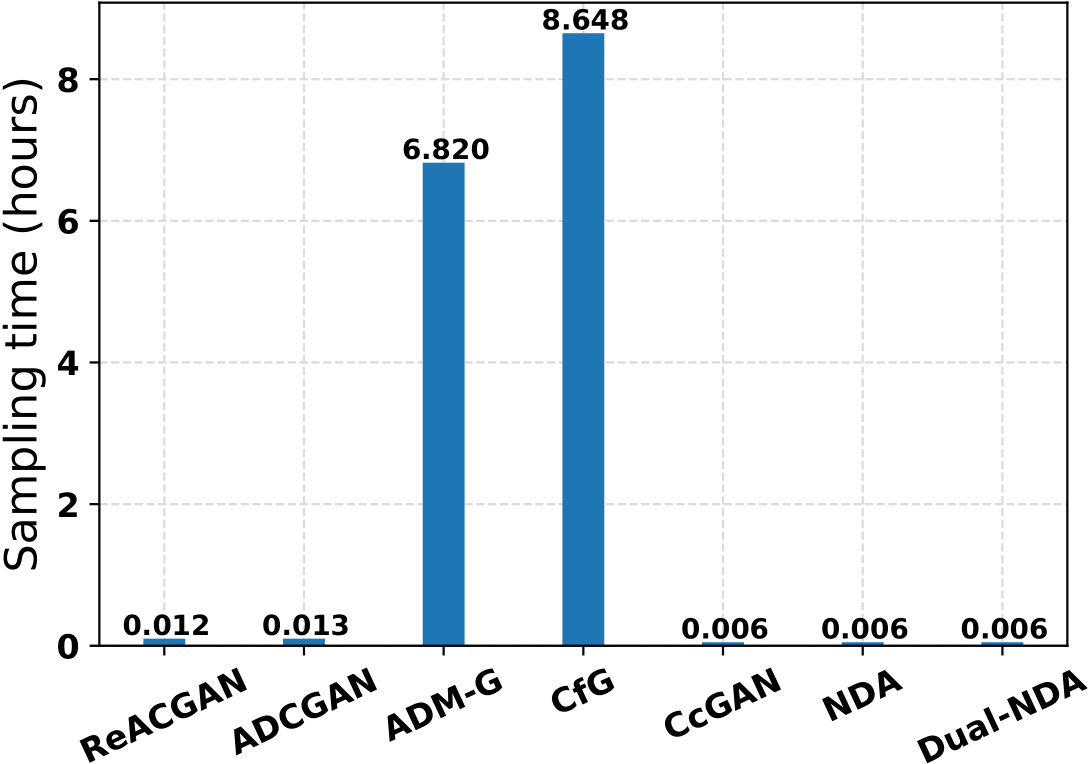}
	\caption{Sampling time comparison on Steering Angle (64$\times$64). Each method is asked to generate 50 fake images for each of the 2000 distinct evaluation angles.}\label{fig:sample_time_SA64}
\end{figure*}

\end{document}